\documentclass[10pt,twocolumn,letterpaper]{article}

\usepackage{iccv}
\usepackage{times}
\usepackage{epsfig}
\usepackage{graphicx}
\usepackage{amsmath}
\usepackage{amssymb}
\usepackage{booktabs}
\usepackage{multirow}
\usepackage{float}

\usepackage[pagebackref=true,breaklinks=true,letterpaper=true,colorlinks,bookmarks=false]{hyperref}

\iccvfinalcopy 

\begin{document}

\title{Revisiting Scene Text Recognition: A Data Perspective}

\author{Qing Jiang , Jiapeng Wang , Dezhi Peng , Chongyu Liu , Lianwen Jin$^{*}$ \\
South China University of Technology\\
{\tt\small mountchicken@outlook.com ,  scutjpwang@foxmail.com, eedzpeng@mail.scut.edu.cn} \\
{\tt\small liuchongyu1996@gmail.com ,  eelwjin@scut.edu.cn} \\
\url{https://union14m.github.io/}
}
\maketitle
\begin{abstract}
  This paper aims to re-assess scene text recognition (STR) from a data-oriented perspective. We begin by revisiting the six commonly used benchmarks in STR and observe a trend of performance saturation, whereby only 2.91\% of the benchmark images cannot be accurately recognized by an ensemble of 13 representative models. While these results are impressive and suggest that STR could be considered solved, however, we argue that this is primarily due to the less challenging nature of the common benchmarks, thus concealing the underlying issues that STR faces. To this end, we consolidate a large-scale real STR dataset, namely Union14M, which comprises 4 million labeled images and 10 million unlabeled images, to assess the performance of STR models in more complex real-world scenarios. Our experiments demonstrate that the 13 models can only achieve an average accuracy of 66.53\% on the 4 million labeled images, indicating that
  STR still faces numerous challenges in the real world. By analyzing the error patterns of the 13 models, we
  identify seven open challenges in STR and develop a challenge-driven benchmark consisting of eight distinct subsets to facilitate further progress in the field. Our exploration demonstrates that STR is far from being solved and leveraging data may be a promising solution. 
  In this regard, we find that utilizing the 10 million unlabeled images through
  self-supervised pre-training can significantly improve the robustness
  of STR model in real-world scenarios and leads to state-of-the-art performance. Code and dataset is available at \url{https://github.com/Mountchicken/Union14M}.
\end{abstract}

\section{Introduction}
\label{sec:intro}
\begin{figure}[t]
  \centering
  \includegraphics[scale=0.35]{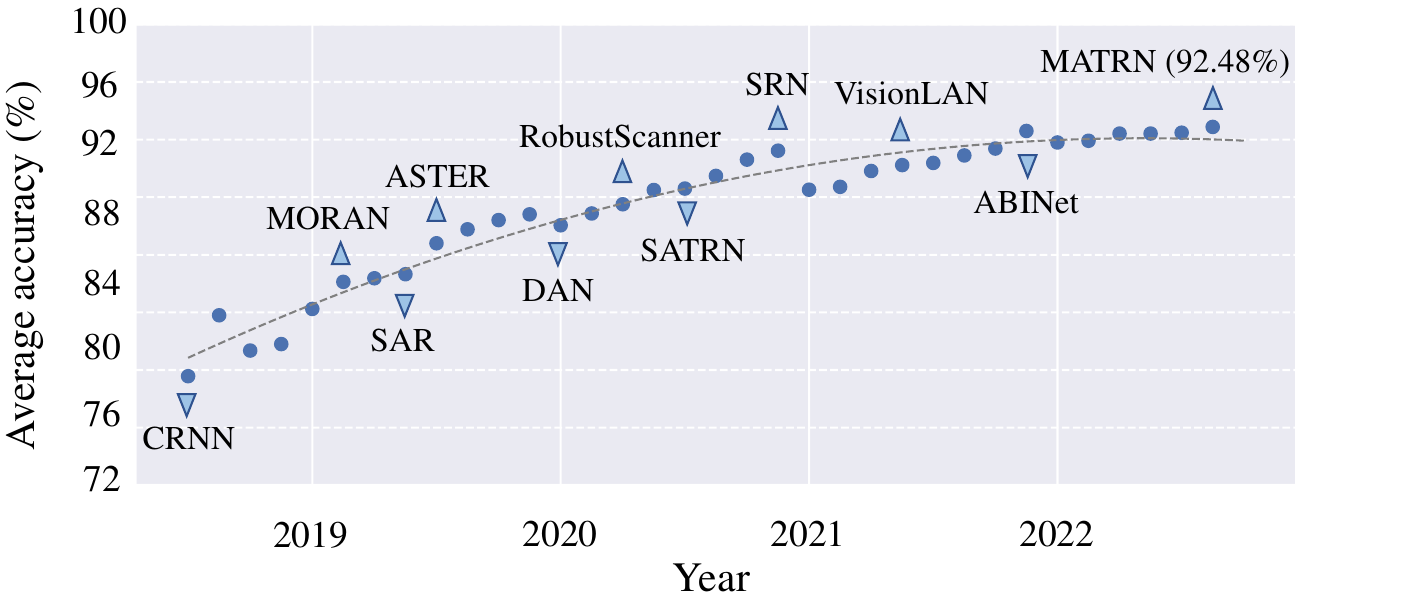}
  \caption{Average accuracy of STR models on six commonly used benchmarks as reported in their original papers. Models are 
  trained with synthetic data.
  }
  \label{fig:1}
  \vspace{-1.5em}
\end{figure}

The success of deep learning in visual recognition tasks
heavily depends on expansive labeled data. A widely used paradigm
 \cite{baek2019wrong, du2022svtr, fang2021read, lu2021master, yu2020towards} in STR is training models on
large-scale synthetic datasets  \cite{jaderberg2014synthetic, jaderberg2016reading, gupta2016synthetic, yim2021synthtiger}
and evaluating on six real benchmarks  \cite{risnumawan2014robust,phan2013recognizing,wang2011end,mishra2012scene,karatzas2013icdar,karatzas2015icdar}.
Promisingly, current progress in STR has exhibited a trend of accuracy
saturation (depicted in Fig. \ref{fig:1}). The challenges in the common benchmarks seem “solved”,
suggested by the narrow scope for improvement, and the slowdown step of performance
gain in recent SOTAs. This phenomenon inspires us to raise questions
of 1) \textit{whether the common benchmarks remain sufficient to promote future progress,}
and 2) \textit{whether this accuracy saturation implies that STR is solved.}

\begin{figure}[t]
  \centering
  \includegraphics[scale=0.36]{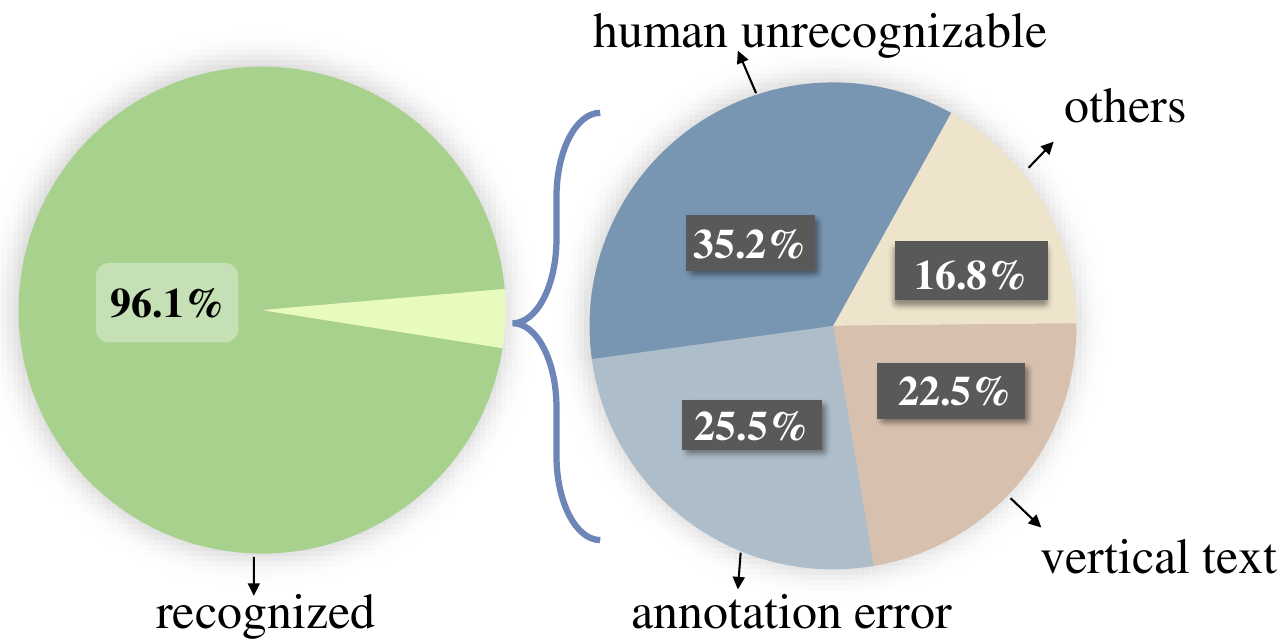}
  \caption{Error analysis on the six STR benchmarks.}
  \label{fig:2}
  \vspace{-1.em}
\end{figure}

For the first question, we start by selecting 13 representative models (listed in Tab. \ref{tab:3}), including CTC-based  \cite{7801919,du2022svtr},
attention-based  \cite{shi2018aster,luo2019moran,li2019show,lee2020recognizing,sheng2019nrtr,wang2020decoupled,yue2020robustscanner},
and language model-based  \cite{yu2020towards, fang2021read, wang2021two, na2022multi} models.
We then evaluate their performance on the six STR benchmarks to find their joint errors. As
depicted in Fig. \ref{fig:2}, only 3.9$\%$ (298 images) of the
total 7672 benchmark images can not be correctly
recognized by any of the 13 models, among which 25.5\% of the images are incorrectly
annotated, and 35.2\% images are barely recognizable
(human unrecognizable samples, shown in Appendix \ref{unrecog_sample}.). This suggests that there
might be a maximum of 2.91$\%$ (222 images) and a
minimum of 1.53\% (117 images, excluding human unrecognizable samples) scope for
accuracy improvement. Therefore, the common benchmarks give limited insight into
future STR research.

\begin{figure}[t]
  \centering
  \includegraphics[scale=0.49]{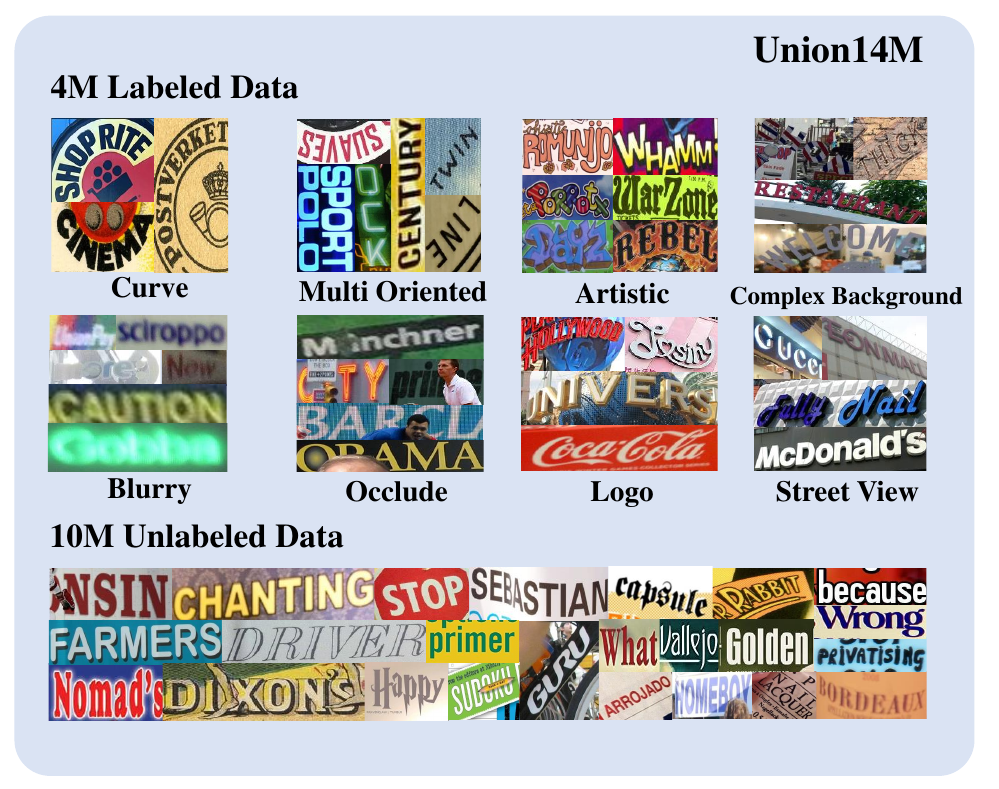}
  \caption{An Overview of Union14M which is used to analyze STR models in real-world scenarios. Union14M contains 4M labeled images (Union14M-L)
    and 10M unlabeled images (Union14M-U), which cover a wide range of real-world scenarios
    with intense diversity and complexity.}
  \label{fig:3}
  \vspace{-1.5em}
\end{figure}

The accuracy saturation in common benchmarks can obscure challenges that STR models
still face. Therefore, to bring more
profound insights beyond these benchmarks and to benefit real-world STR applications, we consolidate a large-scale real
dataset, namely Union14M (Fig. \ref{fig:3}), to carefully analyze the performance of STR models in a broader
range of real-world scenarios. Union14M consists of 4
million \textbf{L}abeled images (Union14M-L) and 10 million \textbf{U}nlabeled
images (Union14M-U), obtained from 17 publicly available datasets. Hence, it can
be considered as a comprehensive representation of text in the real world.


Equipped with Union14M-L, we conducted a quantitative evaluation on the
aforementioned 13 STR models from multiple perspectives, thereby uncovering challenges that remain in STR. Our initial
observation is related to the synthetic-to-real training paradigm. We
discover that models trained on synthetic data perform poorly on Union14M-L,
with an average accuracy of only 66.53\%, despite achieving an average accuracy
of 87.03\% on commonly used benchmarks. This result indicates that such a paradigm is not compatible with more complex real-world settings. Subsequently,
we analyze the error patterns of the 13 models and find that they are still less
robust to four existing challenges, namely \textit{curve text},
\textit{multi-oriented text}, \textit{contextless text}, and \textit{artistic text}.
Furthermore, we identify three additional challenges that are prevalent in the
real world but have received less attention in the STR community, namely \textit{multi-words text}, \textit{salient text},
and \textit{incomplete text}.

To enable more thorough evaluations of STR models in real-world scenarios
and to encourage future research on the seven aforementioned challenges,
we construct a challenge-driven benchmark, which comprises eight
subsets with 400,000 generic samples and 9,383 challenge-specific samples sourced
from Union14M-L. Extensive baseline experiments are conducted on this new
benchmark and we find that
despite utilizing real data for training, the current SOTA model can only achieve an average accuracy of 74.6\%. This indicates that STR still faces numerous
challenges in the real world and also answers the second question that STR is
far from being solved.

Essentially, we infer that the sub-optimal performance of STR models in the real world can be
attributed to data problems, e.g., the lack of sufficient real labeled data for
training. To solve STR from a data perspective, we propose a solution of utilizing unlabeled data. Specifically, we investigate a Vision Transformer-based  \cite{dosovitskiy2020image} STR model (Fig. \ref{fig:5}), which can leverage the 10M unlabeled images in Union14M-U through self-supervised pre-training. The pre-trained ViT model exhibits powerful textual representation capabilities, and after fine-tuning on real labeled data, it achieves SOTA performance on both six common benchmarks and the proposed challenge-driven benchmark. Our contributions are
summarized as follows:

\begin{itemize}
  \setlength{\itemsep}{0pt}
  \setlength{\parsep}{0pt}
  \setlength{\parskip}{0pt}
  \item We analyze STR from a data perspective and arrive at two macro findings. Firstly, the
        common benchmarks are insufficient in presenting adequate challenges for advancing the field of STR. Secondly, despite significant progress, STR models still struggle to perform well in real-world scenarios. It is safe to say that STR is still far from being solved.
  \item We consolidate a large-scale real STR dataset to    
         investigate the
        performance of STR models in the real world. Through quantitative analysis, we reveal
        that current STR models fail to address seven open challenges. Therefore,
        we propose a challenge-driven benchmark to facilitate future comprehensive and in-depth studies in the field of STR.
  \item We exploit the potential of unlabeled data and observe that
  they can lead to significant performance gains through self-supervised pre-training, offering a practical solution for STR in the real world.
  
\end{itemize}

\section{Related Works}

\subsection{Data Analysis in STR}
In scene text recognition, some works have been proposed to analyze several data issues. For instance, Baek \textit{et al.} \cite{baek2019wrong}
point out the inconsistency between the training data and benchmarks in STR approaches.
They also conduct a comprehensive analysis on the
common benchmarks and find that 7.5\% of the images can not be recognized by their proposed
four-stage framework. In this work, we further refined it to 3.9\% by
using 13 distinctive STR models. Baek \textit{et al.} \cite{baek2021if} explored
the impact of real data on the performance of STR models, in which they found that training
on fewer real data can lead to better performance than training on synthetic data, and
several recent works  \cite{yang2022reading,bautista2022scene,singh2021textocr} have confirmed this finding by using real data for training. We also observe in our subsequent experiments that training models on real data can improve their generalization ability, which is essential for real-world STR applications.

\subsection{Data Shift in STR}
Scene text recognition is a fine-grained task that requires
extensive amounts of training data. In the early time, due to the lack of
sufficient real annotated data, STR models were
trained on large-scale synthetic datasets, e.g., MJ  \cite{jaderberg2014synthetic, jaderberg2016reading} and ST  \cite{
gupta2016synthetic}.This training paradigm still prevails today as state-of-the-art methods  \cite{fang2021read, lee2020recognizing, na2022multi}
continuously yield better performance on the common benchmarks. Nevertheless,
models trained on synthetic data might suffer from
generalization problems, due to the large domain gap  \cite{zhang2019sequence,baek2021if}
between synthetic data and real-world circumstances.

Meanwhile, a few annotated real datasets have emerged in recent years  \cite{chng2019icdar2019,
  krylov2021open,singh2021textocr,nayef2019icdar2019, long2022towards}. Serval recent works
have endeavored to consolidate these datasets. For instance, the OOV \cite{garcia2023out} dataset
is a consolidation of seven real datasets and is employed to investigate the out-of-vocabulary \cite{wan2020vocabulary}
problem. Baek \textit{et al.} \cite{baek2021if}, Yang \textit{et al.} \cite{yang2022reading}, and 
Darwin \textit{et al.} \cite{bautista2022scene} use different amounts of real datasets to
construct the training set respectively, and achieved better results
than training on synthetic data. In this work, our aim is to analyze the performance
of STR models in the real world and the challenges they confront. Therefore, we
consolidate Union14M with more real datasets, thus it can be used as a
real-world mapping for our analysis.

\subsection{Benchmarks in STR}
In STR, there are six commonly used benchmarks, including regular text benchmarks: IC13 \cite{karatzas2013icdar},
IIIT \cite{mishra2012scene}, SVT \cite{wang2011end} and irregular text benchmarks: IC15 \cite{karatzas2015icdar},
SVTP \cite{phan2013recognizing}, CUTE \cite{risnumawan2014robust}. Some recent
works \cite{yang2022reading,bautista2022scene} attempt to use alternative
benchmarks \cite{veit2016coco, liu2019curved, ch2017total, chng2019icdar2019}
for evaluation, and they also observe performance degradation
on these benchmarks compared to the six benchmarks. This suggests that there exists challenges
that exceed the scope of common benchmarks and an in-depth analysis is necessary.

\section{Preliminary: A Real Dataset for Analysis}
As previously discussed, the six STR benchmarks have almost reached a point of 
saturation, and can be insufficient to facilitate our analysis across a
broader spectrum of 
real-world scenarios. Hence, we consolidate a large-scale real STR dataset denoted as
Union14M, comprising 4 million labeled images (Union14M-L) and 10 million
unlabeled images (Union14M-U), to support our subsequent analysis.

\subsection{Dataset Consolidation}
\label{sec:dataset}
\textbf{Union14M-L: 4M labeled images. }Our data collection strategy is driven by
the primary objective of encompassing a broad range of real scenarios. To this
end, we collect labeled images from 14 publicly available datasets (Tab. \ref{tab:1})
to compose Union14M-L.
These datasets exhibit diverse properties. For instance, ArT \cite{chng2019icdar2019} dataset is focused on curved
text; ReCTS \cite{zhang2019icdar}, RCTW \cite{shi2017icdar2017}, LSVT \cite{sun2019chinese, sun2019icdar},
KAIST \cite{jung2011touch}, NEOCR \cite{nagy2011neocr} and IIIT-ILST \cite{mathew2017benchmarking} datasets are 
designed for street views from different countries; MTWI \cite{he2018icpr2018} is sourced from web pages and contains scene text images; COCOTextV2 \cite{veit2016coco} contains plenty of
low-resolution text images as well as vertical text images; IntelOCR \cite{krylov2021open}, TextOCR \cite{singh2021textocr}
and HierText \cite{long2022towards} are all derived from OpenImages \cite{kuznetsova2020open},
which is a vast dataset with nine million images covering an extensive range of real scenes.
The consolidation of the 14 datasets can be viewed as a mapping of the real world,
enabling our analysis to be oriented toward real-world scenarios.

\begin{table}[t]
  \centering
  \caption{Composition of Union14M. $^{\dagger}$ denotes that the dataset overlaps with current benchmarks. $^{\ddagger}$ denotes those datasets overlap with each other. \#Original denotes the number of text instances in the original dataset.}
  \resizebox{\columnwidth}{!}{%
    \begin{tabular}{cccccc}
      \toprule
      \multicolumn{1}{l}{}         & Dataset                                          & Year & \#Original & \#Refined & Lang.  \\
      \midrule
      \multirow{14}{*}{Union14M-L} & KAIST \cite{jung2011touch}                  & 2011 & 6K        & 2K         & EN, KR \\
                                   & NEOCR \cite{nagy2011neocr}                        & 2011 & 5K         & 3K          & EN     \\
                                   & Uber-Text \cite{zhang2017uber}                    & 2017 & 209K       & 208K        & EN     \\
                                   & RCTW \cite{shi2017icdar2017}                & 2017 & 44K        & 7K          & EN, CH \\
                                   & IIIT-ILST \cite{mathew2017benchmarking}     & 2017 & 6K        & 2K         & EN, IN \\
                                   & MTWI \cite{he2018icpr2018}                  & 2018 & 139K       & 53K         & EN, CN \\
                                   & COCOTextV2 \cite{veit2016coco}                    & 2018 & 201K       & 73K         & EN     \\
                                   & LSVT \cite{sun2019chinese, sun2019icdar}    & 2019 & 382K       & 38K         & EN, CN \\
                                   & MLT19 \cite{nayef2019icdar2019}             & 2019 & 89K        & 56K         & Multi  \\
                                   & ReCTS \cite{zhang2019icdar}                 & 2019 & 109K       & 25K         & EN, CN \\
                                   & ArT$^{\dagger}$ \cite{chng2019icdar2019}        & 2019 & 50K        & 35K         & EN, CN \\
                                   & IntelOCR$^{\ddagger}$ \cite{krylov2021open}       & 2021 & 2.57M      & 2.01M       & EN     \\
                                   & TextOCR$^{\ddagger}$] \cite{singh2021textocr}     & 2021 & 822K       & 586K        & EN     \\
                                   & HierText$^{\ddagger}$ \cite{long2022towards}      & 2022 & 1.2M       & 945K        & EN     \\
      \midrule
      \multirow{3}{*}{Union14M-U}  & Book32 \cite{iwana2016judging}                    & 2016 & -          & 2.7M        & -      \\                   
                                   & CC \cite{sharma2018conceptual}                    & 2018 & -          & 5.6M        & -      \\
                                   & OpenImages$^{\ddagger}$ \cite{kuznetsova2020open} & 2020 & -          & 2.3M        & -      \\
      \bottomrule
    \end{tabular}}
  \label{tab:1}
  \vspace{-1.5em}
\end{table}

Nevertheless, the simple concatenation of these 14 datasets is sub-optimal due
to different annotation formats and the existence of duplicate, Non-Latin, and corrupted samples. Hence,
we adopt the following strategies to refine.
\begin{itemize}
  \setlength{\itemsep}{0pt}
  \setlength{\parsep}{0pt}
  \setlength{\parskip}{0pt}
  \item \textbf{Crop text instances.} Most datasets provide polygon
  annotations for text instances, and directly using the polygon for cropping
  is an intuitive choice. However, we conjecture this could be sub-optimal.
  Instead, we use the minimum axis-aligned rectangle for cropping, which can
  result in additional background noise for cropped text instances. This
  cropping strategy essentially serves as a form of regularization, as it introduces
  challenging samples (i.e., those with more background noise) that enhance the
  robustness of the recognizer. This is beneficial in an end-to-end system, as the recognizer
  can be less dependent on the performance of the detector, and also allows us to
  focus our analysis on the performance of the recognizer. We validate this conjecture about
  cropping methods in Appendix \ref{crop}.
  \item \textbf{Exclude duplicate samples.} We first remove duplicate
  samples between Union14M-L and the common benchmarks. Next, we remove duplicate samples
  among the 14 datasets. For instance, HierText, TextOCR, and IntelOCR 
  are duplicated with each other since they are all annotated from 
  OpenImages \cite{kuznetsova2020open}. We choose
  HierText as reference, and remove duplicated samples from the remaining two datasets.
  \item \textbf{Remove Non-Latin and ignored samples.} In this work, we
  focus on Latin characters which are widely employed and possess a large
  amount of data. Consequently, We only retained samples composed of
  letters, numbers, and symbols. We also remove samples that are
  annotated as ignored.
\end{itemize}
\vspace{-0.5em}

\textbf{Union14M-U: 10M unlabeled images. }Self-supervised learning has enabled
substantial development in
computer vision \cite{xie2022simmim,he2022masked,he2020momentum,chen2020simple}, and several related works
have also emerged in the field of STR \cite{luo2022siman, yang2022reading, lyu2022maskocr, aberdam2021sequence}.
The optimal solution to improve the performance of STR in real-world scenarios is to
utilize more data for training. However, labeling text images is both costly and
time-intensive, given that it involves annotating
sequences and needs specialized language expertise. Therefore, it would be desirable to
investigate the potential of utilizing unlabeled data via self-supervised learning for STR.
To this end, we collect 10M unlabeled images from three large datasets, including
Book32 \cite{iwana2016judging}, OpenImages \cite{kuznetsova2020open} and
Conceptual Captions (CC) \cite{sharma2018conceptual} dataset. To obtain high-quality text instances, we
adopt a different collection method than previous works \cite{yang2022reading, baek2021if}. We use three
text detectors \cite{zhou2017east, liao2022real, liu2019omnidirectional} and an IoU voting 
mechanism to get text instances (detailed in Appendix \ref{construct_union14m}). The unlabeled images collected from OpenImages 
are also de-duplicated with the labeled images in Union14M-L.
\begin{table}[]
  \caption{Statistics of Union14M and synthetic datasets MJ \cite{jaderberg2014synthetic, jaderberg2016reading} and ST \cite{
    gupta2016synthetic}. Vertical instances are text images with a height that is at least twice their width and with more than one text character.}
  \resizebox{\columnwidth}{!}{%
    \begin{tabular}{lccc}
      \toprule
      Dataset    & \multicolumn{1}{l}{\# Instances} & \multicolumn{1}{l}{\# Vocabularies} & \multicolumn{1}{l}{\# Vertical Instances} \\ 
      \midrule
      MJ+ST      & 17M                              & 384K                               & 7K                                        \\
      Union14M-L & 4M                               & 707K                               & 110K                                      \\
      Union14M-U & 10M                              & -                                  & 39K                                       \\
      \bottomrule
    \end{tabular}%
  } 
  \label{tab:2}
  \vspace{-2.0em}
\end{table}
\subsection{Characteristics of Real-World Data}
\textbf{Diverse text styles. }As shown in Fig. \ref{fig:3}, Union14M covers text images from a variety of real scenes. Real-world text images exhibit diverse layouts, e.g.,
curve, tilted and vertical, as well as challenging distractions, including
blurring, complex background and occlusion, and also various real-world applications
of scene text, such as street scenes and logos. Notably, Union14M contains a large
number of vertical text instances (last column in Tab. \ref{tab:2}), which are
common in real world, yet are rare in synthetic datasets.

\textbf{Large vocabularies. }Text used in synthetic datasets are obtained from
commonly used corpus. However, in real-world scenarios, there are plenty of text variations that are
not covered by corpus, such as random combinations of alphanumeric characters and symbols, for instance,
license plates, or multilingual alphabetical combinations like Chinese Pinyin. In Tab. \ref{tab:2},
we show that the number of vocabularies in Union14M-L is nearly twice as larger as that of
synthetic datasets, demonstrating that Union14M-L can encompass a broader spectrum of real-world situations
and thus can hold our further analysis.

\section{Analysis of STR in Real World}
In this section, we utilize the vast nature of Union14M-L to
conduct a comprehensive analysis of the performance of 13 STR models. The objective
of this analysis is to evaluate the robustness of STR models
against numerous real-world challenges, identify existing challenges, and
stimulate future research advances.

\begin{table}
  \caption{We use 13 publicly available models for evaluation. Acc-CB represents the average
    accuracy on six commonly used benchmarks. Acc-UL represents the accuracy on
    all Union14M-L data.}
  \resizebox{\columnwidth}{!}{%
    \begin{tabular}{lcccc}
      \toprule
      Method                                   & Type & Venue    & Acc-CB & Acc-UL                           \\
      \midrule
      CRNN \cite{7801919}                       & CTC & TPAMI'17 & 78.14 & 57.96 (\textcolor{red}{-20.18}) \\
      SVTR \cite{du2022svtr}                    & CTC & IJCAI'22  & 90.00 & 69.46 (\textcolor{red}{-20.54}) \\
      \midrule
      MORAN \cite{luo2019moran}                 & Att. & PR'17     & 80.61 & 57.73 (\textcolor{red}{-22.88}) \\
      ASTER \cite{shi2018aster}                 & Att. & TPAMI'19 & 84.98 & 63.30 (\textcolor{red}{-21.68}) \\
      NRTR \cite{sheng2019nrtr}                 & Att. & ICDAR'19  & 86.82 & 66.96 (\textcolor{red}{-19.86}) \\
      SAR \cite{li2019show}                     & Att. & AAAI'19   & 88.07 & 68.07 (\textcolor{red}{-20.00}) \\
      DAN \cite{wang2020decoupled}              & Att. & AAAI'20  & 83.96 & 64.16 (\textcolor{red}{-19.80}) \\ 
      SATRN \cite{lee2020recognizing}           & Att. & CVPRW'20  & 91.36 & 72.09 (\textcolor{red}{-19.27}) \\
      RobustScanner \cite{yue2020robustscanner} & Att. & ECCV'20  & 87.63 & 67.63 (\textcolor{red}{-20.00}) \\
      \midrule
      SRN \cite{yu2020towards}                  & LM & CVPR'20   & 86.51 & 65.71 (\textcolor{red}{-20.80}) \\
      ABINet \cite{fang2021read}                & LM & CVPR'21   & 91.97 & 70.73 (\textcolor{red}{-21.24}) \\
      VisionLAN \cite{wang2021two}              & LM & ICCV'21   & 88.96 & 69.60 (\textcolor{red}{-19.36}) \\ 
      MATRN \cite{na2022multi}                  & LM & ECCV'22   & 92.48 & 71.49 (\textcolor{red}{-20.99}) \\
      \bottomrule
    \end{tabular}}
  \label{tab:3}
  \vspace{-1.5em}
\end{table}

\subsection{Overall Performance Evaluation}
\begin{figure*}[t]
  \centering
  \includegraphics[scale=0.44]{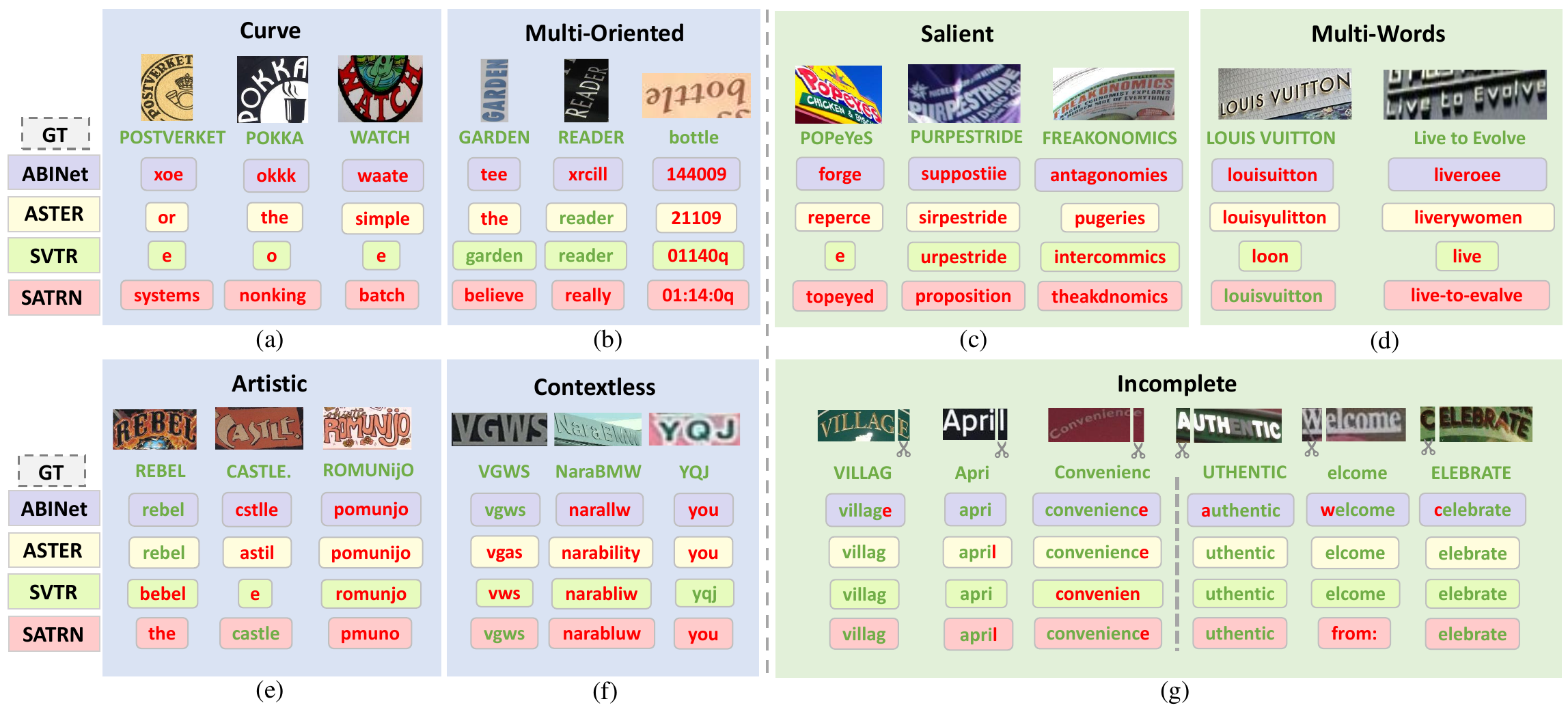}
  \caption{Error analysis of the 13 STR models. We select four representative
    models and show their prediction results (Text in green represents correct
    prediction and red text vice versa). Blocks in blue (a, b, e, f) represent four
    unsolved challenges, and blocks in green (c, d, g) represent three additional
    challenges that are rarely discussed. Best viewed in color.}
  \label{fig:4}
  \vspace{-0.5em}
\end{figure*}

We begin by selecting 13 representative models
trained on synthetic datasets to evaluate on Union14M-L. As shown in Tab. \ref{tab:3}, compare to the performance on common benchmarks, their performance degradation on Union14M-L is significant, with
an average accuracy drop of 20.50$\%$. This result suggests that  models trained on synthetic data can not well generalize to more
complicated real-world scenarios. Conversely, it also suggests that  Union14M-L features challenges that are not covered by common benchmarks and worth a deeper investigation.

\subsection{Challenge Mining}
\label{sec:problems_mining}
To identify the joint errors made by the 13 models, we assign each sample in
Union14M-L with a difficulty score based on the number of correct predictions (detailed in Appendix \ref{assignment}).
We focus on the hard samples that the majority of models fail to make
correct predictions, and we summarize four challenges that haven't been adequately solved (left side of Fig. \ref{fig:4}). Furthermore, we introduce three additional
challenges that are common in the real world, yet are seldom discussed in
previous works (right side of Fig. \ref{fig:4}).

\subsubsection{Unsolved Challenges}
\label{sec:unsolved_problems}
\textbf{Curve Text.} Curve text recognition has gained considerable attention in recent
years, with two mainstream approaches: one that relies on
rectification \cite{luo2019moran, shi2018aster} and
the other that employs 2D attention mechanism \cite{lee2020recognizing,li2019show,yue2020robustscanner}.
Both approaches yield promising results on the curve text benchmark
CUTE \cite{risnumawan2014robust}. However, the proportion of curve text in
the CUTE benchmark is limited, and the extent of curvature is minor. For highly curved
texts as shown in Fig. \ref{fig:4}a, current methods still exhibit limited performance.

\textbf{Multi-Oriented Text.} Text can appear on the surface of any object in
any orientation, including vertical, tilted, or reversed cases (Fig. \ref{fig:4}b). Multi-oriented
text is common in real-world scenarios, such as vertical text on billboards,
tilted text due to the shooting angle of the camera, and reversed text due to mirror
reflection. However, this problem is overlooked in most STR methods with
a strong assumption that text images are nearly horizontal. They followed a similar
procedure of scaling the height of text images to a small size (e.g., 32 pixels), and then
scaling the width while keeping the ratio unchanged, causing vertical or tilted images to collapse
in height and consequently impeding recognition.

\textbf{Artistic Text.} In contrast to printed text, artistic text is designed by artists
or professional designers with diverse text fonts, text effects, text layouts, and
complex backgrounds. Each instance of the artistic text is potentially unique, making it a
zero-shot or one-shot problem, and may require specifically designed
networks \cite{xie2022toward} for recognition. Nevertheless, due to the lack of artistic text samples in the
synthetic datasets, current models are still less robust to the artistic text shown in Fig. \ref{fig:4}e.

\textbf{Contextless Text.} Contextless text refers to text that has no semantic
meaning and is not in the dictionary. It can be abbreviations or random combinations of letters, digits, and symbols. 
As shown in Fig. \ref{fig:4}f, models may fail to recognize contextless text even when it
has a clear background and minimal distortion. This issue can arise from the
over-introduction of semantic information in both the model design and
dataset corpus, which is also known as vocabulary reliance \cite{wan2020vocabulary, garcia2023out}.
Models will attempt to predict text that appeared in the training
set that follows syntax rules (e.g., mistaking ``YQJ" for ``you" in Fig. \ref{fig:4}f).
This behavior is highly undesirable in applications where reliability is
critical, e.g., license plate recognition, invoice recognition, and card ID recognition,
where most of the text are contextless and their misrecognition can lead to enormous
security risks and property damages.

\subsubsection{Additional Challenges}

\textbf{Salient Text. }Salient text refers to the presence of extra characters
that coexist with the primary characters of interest in a text image (Fig. \ref{fig:4}c).
Salient text can be inadvertently introduced in end-to-end text recognition when text
instances of different sizes are adjacent or overlapping with each other. This problem
has been discussed in the text detection stage. For instance, Liao \textit{et al.} \cite{liao2020mask} propose to
use a hard ROI masking strategy to eliminate the interference of extra characters.
Nevertheless, when the performance of the detection model is poor, e.g., when it can only output coarse text regions,
it becomes crucial for recognition models to rapidly identify visually important regions. However, as shown in Fig. \ref{fig:4}c, models
can be confused by additional characters and fail to recognize the primary text.

\textbf{Multi-Words Text. }Text contains rich semantic information that aids in
the comprehension of scenes, and sometimes a single word may be insufficient.
In certain cases, the recognition of multiple words simultaneously is required to
fully interpret a text image, such as trademarks and short phrases, as depicted in Fig. \ref{fig:4}d.
However, most STR models are trained on synthetic datasets that comprise a single word per text image, hence
failing to recognize spaces that separate individual words. Moreover, We observe that
models tend to amalgamate multiple words into a single word, discarding or altering
visible characters based on syntax rules (e.g., ``Live to Evolve" being identified
as ``liveroee" as it reads more like a single word)."

\textbf{Incomplete Text. }Text images can be incomplete, with missing characters
due to occlusion or inaccurate detection boxes that truncate the text. In
Fig. \ref{fig:4}g, when a text image is
cropped with the first or the last letter, models may produce completed predictions, 
even though the missing letter is invisible. Moreover, we observe that
this behavior occurs more frequently in language models (Sec. \ref{sec:experiment_results}) that rely heavily on linguistic
priors. This feature may reduce the reliability of models in text analysis applications.
For instance, a fragmented text image with ``ight'' written on it may be completed
as ``might'' or ``light'', while it would be optimal for the recognition model to
output what it actually sees, i.e. ``ight'', thus allowing anomaly detection.
Therefore, it is crucial to thoroughly evaluate the performance
of the automatic completion feature and consider the potential impact on
downstream applications.

\begin{table}[]
  \centering
  \caption{Dataset partion of Union14M-L. Union14M-Benchmark is a split from Union14M-L.}
  \resizebox{0.75\columnwidth}{!}{%
    \begin{tabular}{lcccl}
      \toprule
      \multicolumn{1}{c}{\multirow{2}{*}{Dataset}} & \multicolumn{3}{c}{\#Images}                       \\ \cline{2-4}
      \multicolumn{1}{c}{}                         & Train                        & Val     & Benchmark \\ \hline
      Union14M-L                                   & 3,230,742                    & 400,000 & 409,383   \\
      \midrule
      General                                      & -                            & -       & 400,000   \\
      Artistic                                     & -                            & -       & 900       \\
      Curve                                        & -                            & -       & 2426      \\
      Multi-Orinted                                & -                            & -       & 1369      \\
      Multi-Words                                  & -                            & -       & 829       \\
      Salient                                      & -                            & -       & 1585      \\
      Incomplete                                   & -                            & -       & 1495      \\
      Contextless                                  & -                            & -       & 779       \\
      \bottomrule
    \end{tabular}}
  \label{tab:4}
  \vspace{-1.0em}
\end{table}

\section{A Challenge-Driven Benchmark}
To facilitate the evaluation of STR models in more comprehensive real-world
scenarios and to support future research on the aforementioned seven challenges,
we construct a challenge-driven benchmark, namely Union14M-Benchmark.
It consists of eight subsets and a total of 409,393 images with both complexity
and versatility.

\subsection{Benchmark Construction}
\textbf{Challenge-specific subsets. }We collected subsets for
each of the seven challenges presented in Sec. \ref{sec:problems_mining}. Candidate
images are manually selected from Union14M-L based on some reference samples of
these seven text types, except for the incomplete text. For the incomplete text subset,
we sample 1,495 images that the majority of the 13 models can make
correct predictions from Union14M-L since we aim to investigate the auto-completion feature of STR models and therefore we shall not introduce other factors that might lead to false recognition.  
Then we randomly crop out either the first or the last letter of the text image. To ensure that there are no duplicate images between Union14M-L and the proposed benchmark, we counted the remaining samples in Union14M-L
that have the same text label as the benchmark images, and then we manually reviewed each sample to remove the duplicate images in Union14M-L.

\textbf{General subset.} In addition to these seven specific challenges,
STR poses several other difficulties, such as blurring, chromatic distortion \cite{zhang2021spin}, and complex background \cite{luo2021separating}. Therefore, to enhance the diversity of this benchmark, we also construct a general subset with
400,000 images sampled from Union14M-L.

We also emphasize the significance of the validation set. It follows the same construction methodology as the general subset, which also includes 400,000 samples. The statistics are shown in Tab. \ref{tab:4}.

\begin{table*}[!h]
  \renewcommand{\arraystretch}{1.22}
  \caption{Performance (WAICS) of models trained on \textbf{synthetic datasets} (MJ and ST). For the incomplete text subset, we measure the margin of accuracy
    before and after image cropping, which is the lower the better. }
  \resizebox{\linewidth}{!}{
    \begin{tabular}{cccccccccccccccccc}
      \hline
      \multirow{2}{*}{Type}                          & \multirow{2}{*}{Method} & \multicolumn{7}{c}{Common Benchmarks}                                                                                                                                                                                                                                                                                                                     & \multicolumn{8}{c}{Union14M-Benchmark}                                                                                                                                                              & \multicolumn{1}{l}{}           \\ \cline{3-18} 
                                                     &                         & \begin{tabular}[c]{@{}c@{}}IIIT\\ 3000\end{tabular} & \begin{tabular}[c]{@{}c@{}}IC13\\ 1015\end{tabular} & \begin{tabular}[c]{@{}c@{}}SVT\\ 647\end{tabular} & \begin{tabular}[c]{@{}c@{}}IC15\\ 2077\end{tabular} & \begin{tabular}[c]{@{}c@{}}SVTP\\ 645\end{tabular} & \begin{tabular}[c]{@{}c@{}}CUTE\\ 288\end{tabular} & \multicolumn{1}{c|}{Avg}  & Curve & \begin{tabular}[c]{@{}c@{}}Multi-\\ Oriented\end{tabular} & Artistic & Contextless & Salient & \begin{tabular}[c]{@{}c@{}}Multi-\\ Words\end{tabular} & General & \multicolumn{1}{c|}{Avg}  & \multicolumn{1}{l}{Incomplete $\downarrow$} \\ \hline
      \multirow{2}{*}{CTC}                           & CRNN \cite{7801919}                     & 89.7                                                & 88.3                                                & 82.2                                              & 69.3                                                & 67.8                                               & 71.2                                               & \multicolumn{1}{c|}{78.1} & 7.5   & 0.9                                                       & 20.7     & 25.6        & 13.9    & 25.6                                                   & 32.0    & \multicolumn{1}{c|}{18.0} & 6.4                            \\
                                                     & SVTR \cite{du2022svtr}                     & 94.4                                                & 96.3                                                & 91.6                                              & \textbf{84.1}                                                & 85.4                                               & 88.2                                               & \multicolumn{1}{c|}{90.0} & 63.0  & \textbf{32.1}                                                      & 37.9     & 44.2        & 67.5    & 49.1                                                   & 52.8    & \multicolumn{1}{c|}{\textbf{49.5}} & 4.8                            \\ \hline
      \multicolumn{1}{l}{\multirow{7}{*}{Attention}} & MORAN \cite{luo2019moran}                    & 91.0                                               & 91.3                                               & 83.9                                              & 68.4                                                & 73.3                                               & 75.7                                               & \multicolumn{1}{c|}{80.6}     & 8.9   &  0.7                                                     & 29.4      & 20.7        & 17.9    & 23.8                                                   & 35.2    & \multicolumn{1}{c|}{19.5}     & 6.8                               \\
      \multicolumn{1}{l}{}                           & ASTER \cite{shi2018aster}                   & 93.3                                                & 90.8                                                & 90.0                                              & 74.7                                                & 80.2                                               & 80.9                                               & \multicolumn{1}{c|}{85.0} & 34.0  & 10.2                                                      & 27.7     & 33.0        & 48.2    & 27.6                                                   & 39.8    & \multicolumn{1}{c|}{31.5} & 5.8                            \\
      \multicolumn{1}{l}{}                           & NRTR \cite{sheng2019nrtr}                    & 95.2                                                & 94.0                                                & 90.0                                              & 74.1                                                & 79.4                                               & 88.2                                               & \multicolumn{1}{c|}{86.8} & 31.7  & 4.4                                                       & 36.6     & 37.3        & 30.6    & \textbf{54.9}                                                   & 48.0    & \multicolumn{1}{c|}{34.8} & 7.3                            \\
      \multicolumn{1}{l}{}                           & SAR \cite{li2019show}                      & 95.0                                                & 93.7                                                & 89.6                                              & 79.0                                                & 82.2                                               & 88.9                                               & \multicolumn{1}{c|}{88.1} & 44.3  & 7.7                                                       & 42.6     & 44.2        & 44.0    & 51.2                                                   & 50.5    & \multicolumn{1}{c|}{40.6} & \textbf{4.5}                            \\
      \multicolumn{1}{l}{}                           & DAN \cite{wang2020decoupled}                     & 93.4                                                & 92.1                                                & 87.5                                              & 71.6                                                & 78.0                                               & 81.3                                               & \multicolumn{1}{c|}{84.0}     & 26.7  & 1.5                                                      & 35.0      & 40.3        & 36.5    & 42.2                                                   & 42.1    & \multicolumn{1}{c|}{37.4}     & 6.7                               \\
      \multicolumn{1}{l}{}                           & SATRN \cite{lee2020recognizing}                    & 96.1                                                & 95.7                                                & 93.5                                              & \textbf{84.1}                                & 88.5                                               & 90.3                                               & \multicolumn{1}{c|}{91.4} & 51.1  & 15.8                                                      & \textbf{48.0}     & 45.3        & 62.7    & 52.5                                                   & 58.5    & \multicolumn{1}{c|}{47.7} & 5.6                            \\ 
      \multicolumn{1}{l}{}                           & RobustScanner \cite{yue2020robustscanner}           & 95.1                                                & 93.1                                                & 89.2                                              & 77.8                                                & 80.3                                               & 90.3                                               & \multicolumn{1}{c|}{87.6} & 43.6  & 7.9                                                       & 41.2     & 42.6        & 44.9    & 46.9                                                   & 39.5    & \multicolumn{1}{c|}{38.1} & \textbf{4.5}                            \\ \hline
      \multirow{4}{*}{LM}                            & SRN \cite{yu2020towards}                     & 91.5                                                & 93.9                                                & 88.9                                              & 76.0                                                & 84.0                                               & 84.8                                               & \multicolumn{1}{c|}{86.5} & \textbf{63.4}  & 25.3                                                      & 34.1     & 28.7        & 56.5    & 26.7                                                   & 46.3    & \multicolumn{1}{c|}{39.6} & 7.6                            \\                                                    
                                                     & ABINet \cite{fang2021read}                  & 95.7                                                & 95.7                                                & 94.6                                              & 85.1                                                & 90.4                                               & 90.3                                               & \multicolumn{1}{c|}{92.0} & 59.5  & 12.7                                                      & 43.3     & 38.3        & 62.0    & 50.8                                                   & 55.6    & \multicolumn{1}{c|}{46.0} & 17.9                           \\
                                                     & VisionLAN \cite{wang2021two}               & 95.9                                                & 94.4                                                & 90.7                                              & 80.1                                                & 85.3                                               & 88.9                                               & \multicolumn{1}{c|}{89.2}     & 57.7  & 14.2                                                      & 47.8     & \textbf{48.0}      &  64.0    & 47.9                                                   & 52.1    & \multicolumn{1}{c|}{47.4}     & 6.9                               \\
                                                     & MATRN \cite{na2022multi}                   & \textbf{96.7}                                                & \textbf{95.8}                                                & \textbf{94.9}                 & 82.9                                                & \textbf{90.5}                                               & \textbf{94.1}                               & \multicolumn{1}{c|}{\textbf{92.5}} & \textbf{63.1}  & 13.4                                     & 43.8     & 41.9        & \textbf{66.4}    & 53.2                                                   & \textbf{57.0}    & \multicolumn{1}{c|}{48.4} & 8.2                            \\ \hline
      \end{tabular}}
  \label{tab:5}
\end{table*}

\begin{table*}[!h]
  \renewcommand{\arraystretch}{1.22}
  \caption{Performance (WAICS) of models trained on the training set of \textbf{Union14M-L}. For MAERec,
    S and B represent the use of ViT-Small and ViT-Base as the backbone, respectively. PT denotes pre-training.}
  \resizebox{\linewidth}{!}{
    \begin{tabular}{cccccccccccccccccc}
      \hline
      \multirow{2}{*}{Type}                          & \multirow{2}{*}{Method} & \multicolumn{7}{c}{Common Benchmarks}                                                                                                                                                      & \multicolumn{9}{c}{Union14M-Benchmark}                                                                                                                                                                                                                                                             \\ \cline{3-18} 
                                                     &                         & \begin{tabular}[c]{@{}c@{}}IIIT\\ 3000\end{tabular} & \begin{tabular}[c]{@{}c@{}}IC13\\ 1015\end{tabular} & \begin{tabular}[c]{@{}c@{}}SVT\\ 647\end{tabular} & \begin{tabular}[c]{@{}c@{}}IC15\\ 2077\end{tabular} & \begin{tabular}[c]{@{}c@{}}SVTP\\ 645\end{tabular} & \begin{tabular}[c]{@{}c@{}}CUTE\\ 288\end{tabular}    & \multicolumn{1}{c|}{Avg}               & Curve                  & \begin{tabular}[c]{@{}c@{}}Multi-\\ Oriented\end{tabular}  & Artistic        & Contextless         & Salient                    & \begin{tabular}[c]{@{}c@{}}Multi-\\ Words\end{tabular}                                       & General              & \multicolumn{1}{c|}{Avg}  & \multicolumn{1}{l}{Incomplete $\downarrow$} \\ \hline
  \multirow{2}{*}{CTC}                           & CRNN \cite{7801919}                    & 90.8                                                & 91.8                                                & 83.8                                              & 71.8                                                & 70.4                                                   & 80.9                                     & \multicolumn{1}{c|}{81.6}              & 19.4                      & 4.5                                                     & 34.2            & 44.0                & 16.7                       & 35.7                                     & 60.4                 & \multicolumn{1}{c|}{30.7} & \textbf{0.9}                            \\
                                                  & SVTR \cite{du2022svtr}                    & 95.9                                                & 95.5                                                & 92.4                                              & 83.9                                                & 85.7                                                  & 93.1                                  & \multicolumn{1}{c|}{91.1}              & 72.4                       & 68.2                                                   & 54.1             & 68.0               & 71.4                       & 67.7                                     & 77.0                 & \multicolumn{1}{c|}{68.4} & 2.0                            \\ \hline
  \multicolumn{1}{l}{\multirow{7}{*}{Attention}} & MORAN \cite{luo2019moran}                   & 94.7                                             & 94.3                                               & 89.0                                             &  78.8                                                & 83.4                                                   &  87.2                                   & \multicolumn{1}{c|}{87.9}                & 43.8                       & 12.8                                                   & 47.3            &  55.1               & 45.7                      & 54.6                                      & 44.7                & \multicolumn{1}{c|}{43.4}     &  1.9                              \\
  \multicolumn{1}{l}{}                           & ASTER \cite{shi2018aster}                   & 94.3                                                & 92.6                                               & 88.9                                              & 77.7                                                & 80.5                                                   & 86.5                               & \multicolumn{1}{c|}{86.7}              & 38.4                      & 13.0                                                    & 41.8             & 52.9               & 31.9                       & 49.8                                    & 66.7                 & \multicolumn{1}{c|}{42.1} & 1.3                            \\
  \multicolumn{1}{l}{}                           & NRTR \cite{sheng2019nrtr}                   & 96.2                                                & 96.9                                                & 94.0                                              & 80.9                                                & 84.8                                                   & 92.0                                & \multicolumn{1}{c|}{90.8}              & 49.3                      & 40.6                                                    & 54.3              & 69.6               & 42.9                      & 75.5                                    & 75.2                 & \multicolumn{1}{c|}{58.2}                      & 1.5                            \\
  \multicolumn{1}{l}{}                           & SAR \cite{li2019show}                     & 96.6                                                & 96.0                                                & 92.4                                              & 82.0                                                & 85.7                                                   & 92.7                                   & \multicolumn{1}{c|}{90.9}              & 68.9                        & 56.9                                                  & 60.6             & 73.3               & 60.1                       & 74.6                                    & 76.0                 & \multicolumn{1}{c|}{67.2} & 2.1                            \\                                 
  \multicolumn{1}{l}{}                           & DAN \cite{lee2020recognizing}                     & 95.5                                                & 95.2                                                & 88.6                                              & 78.3                                                & 79.9                                                   & 86.1                          & \multicolumn{1}{c|}{87.3}              & 46.0                       & 22.8                                                   & 49.3             & 61.6               & 44.6                       & 61.2                                   & 67.0                 & \multicolumn{1}{c|}{50.4} & 2.3                               \\
  \multicolumn{1}{l}{}                           & SATRN \cite{wang2020decoupled}                   & 97.0                                                & 97.9                                                & 95.2                                              & 87.1                                                & 91.0                                                   & 96.2                             & \multicolumn{1}{c|}{93.9}              & 74.8                        & 64.7                                                  & 67.1             & 76.1               & 72.2                       & 74.1                                   & 75.8                 & \multicolumn{1}{c|}{72.1} & 0.9                            \\
  \multicolumn{1}{l}{}                           & RobustScanner \cite{yue2020robustscanner}           & 96.8                                                & 95.7                                                & 92.4                                              & 86.4                                                & 83.9                                                   & 93.8                          & \multicolumn{1}{c|}{91.2}              & 66.2                         & 54.2                                                  & 61.4            & 72.7                & 60.1                      & 74.2                                   & 75.7                 & \multicolumn{1}{c|}{66.4} & 1.9                            \\ \hline
  \multicolumn{1}{c}{\multirow{4}{*}{LM}}        & SRN \cite{yu2020towards}                     & 95.5                                                & 94.7                                                & 89.5                                              & 79.1                                                & 83.9                                                   & 91.3                                & \multicolumn{1}{c|}{89.0}              & 49.7                        & 20.0                                                   & 50.7            & 61.0              & 43.9                        & 51.5                                  & 62.7                 & \multicolumn{1}{c|}{48.5} & 2.2                            \\
  \multicolumn{1}{c}{}                           & ABINet \cite{fang2021read}                  & 97.2                                                & 97.2                                                & 95.7                                              & 87.6                                                & 92.1                                                   & 94.4                                   & \multicolumn{1}{c|}{94.0}              & 75.0                          & 61.5                                                 & 65.3             & 71.1               & 72.9                      & 59.1                                 & 79.4                 & \multicolumn{1}{c|}{69.2} & 2.6                            \\
  \multicolumn{1}{c}{}                           & VisionLAN \cite{wang2021two}               & 96.3                                                & 95.1                                                & 91.3                                              & 83.6                                                & 85.4                                                   & 92.4                                   & \multicolumn{1}{c|}{91.3}              & 70.7                        & 57.2                                                   & 56.7             & 63.8               & 67.6                      & 47.3                                & 74.2                 & \multicolumn{1}{c|}{62.5} & 1.3                               \\
  \multicolumn{1}{c}{}                           & MATRN \cite{na2022multi}                   & 98.2                                                & 97.9                                                & 96.9                                              & 88.2                                                & 94.1                                                   & 97.9                                  & \multicolumn{1}{c|}{95.5}              & 80.5                        & 64.7                                                   & 71.1             & 74.8               & 79.4                      & 67.6                                 & 77.9                 & \multicolumn{1}{c|}{74.6} & 1.7                            \\ \hline
  \multicolumn{1}{c}{\multirow{4}{*}{Ours}}      & MAERec-S w/o PT                & 97.4                                                & 97.3                                                & 95.7                                              & 86.7                                                & 91.0                                                   & 96.2                                                  & \multicolumn{1}{c|}{94.1}              & 75.4                        & 66.5                                                   & 66.0             & 76.1               & 72.6                      & 77.0                                 & 80.8                 & \multicolumn{1}{c|}{73.5} & 3.5                            \\
  \multicolumn{1}{c}{}                           & MAERec-S                & 98.0                                                & 97.6                                                & 96.8                                              & 87.1                                                & 93.2                                                   & 97.9                                               & \multicolumn{1}{c|}{95.1}              & 81.4                         & 71.4                                                  & 72.0             & 82.0                & 78.5                     & 82.4                                 & 82.5                 & \multicolumn{1}{c|}{78.6} & 2.7                            \\
  \multicolumn{1}{c}{}                           & MAERec-B w/o PT                     & 97.3                                                & 97.8                                                & 96.6                                              & 87.1                                                & 92.6                                                   & 95.8                                                  & \multicolumn{1}{c|}{94.5}              & 76.5                        & 67.5                                                   & 65.7             & 75.5               & 74.6                      & 77.7                                 & 81.8                 & \multicolumn{1}{c|}{74.2} & 3.2                            \\ 
  \multicolumn{1}{c}{}                           & MAERec-B                & \textbf{98.5}                                                & \textbf{98.1}                                                & \textbf{97.8}                                              & \textbf{89.5}                                                & \textbf{94.4}                                                   & \textbf{98.6}                                              & \multicolumn{1}{c|}{\textbf{96.2}}              &  \textbf{88.8}                      &   \textbf{83.9}                                                  & \textbf{80.0}          & \textbf{85.5}    & \textbf{84.9}  & \textbf{87.5}                & \textbf{85.8} & \multicolumn{1}{c|}{\textbf{85.2}}    & 2.6           \\ \hline
      \end{tabular}}
  \label{tab:6}
  \vspace{-0.5em}
\end{table*}

\section{Experiments and Analysis}

In this section, we benchmark the aforementioned 13 STR models (Tab. \ref{tab:3}) on Union14M-L
to provide more quantitative analysis. In addition, we also introduce a solution for STR from a data perspective by proposing a ViT-based
model \cite{dosovitskiy2020image}, namely \textbf{MAERec} (Sec. \ref{sec:unlabled_data}), which can utilize the 10M unlabeled images in Union14M-U through self-supervised pre-training.

\subsection{Experiment Settings}
\textbf{Training settings. }For the 13 STR models, we use their default
hyperparameters described in the original papers for a fair comparison,
except that the number of the predicted character classes is unified to 91
(including digits, upper and lower case letters, symbols, and space).

\textbf{Metrics. }We use three evaluation metrics: word accuracy (WA),
word accuracy ignoring case (WAIC) and word accuracy ignoring case and symbols
(WAICS, most commonly used). For the incomplete text subset, we measure the margin
of accuracy before and after the letter cropping.

\subsection{Experiment Results}
\label{sec:experiment_results}

\textbf{Real-world data is challenging. }As shown in Tab. \ref{tab:5} and
Tab. \ref{tab:6}, compared to the performance on common benchmarks, models exhibit
an average accuracy degradation of 48.5\% and 33.0\% on Union14M-Benchmark, when
trained on synthetic datasets and Union14M-L respectively. This indicates that
the text images in real-world scenarios is far more complex than the six commonly used
benchmarks.

\textbf{Real-world data is effective. }Models trained on Union14M-L can gain an average
accuracy improvement of 3.9\% and 19.6\% on common benchmarks and Union14M-Benchmark, 
respectively. The large performance boost on Union14M-Benchmark suggests
that synthetic training data can hardly accommodate complex real-world
demands, while using real data for training can largely overcome this generalization
problem. Additionally, the relatively small performance gains on common benchmarks
also imply their saturation.

\textbf{STR is far from being solved. }When trained only on Union14M-L, we observe that
the maximum average accuracy on Union14M-Benchmark (excluding incomplete text subset)
is only 74.6\% (by MATRN \cite{na2022multi} in Tab .\ref{tab:6}). This indicates that STR is far from being solved. Although
relying on large-scale real data can bring a certain performance improvement,
future efforts are still needed.

\textbf{Vocabulary reliance is ubiquitous. }When trained on synthetic datasets, all
models exhibit a large performance drop on incomplete text subset (last column of Tab. \ref{tab:5}).
In particular, we observe that language models have a larger
performance degradation (10.2\% \textit{vs.} 5.6\% in CTC-based
and 5.9\% in attention-based models). We speculate that the performance drop in language models can 
be related to their error correction behavior, i.e., models complete the incomplete
text which is viewed as a character missing error. This problem can be significantly
alleviated when trained on Union14M-L. We attribute
this to the larger vocabulary size in Union14M-L that models will not overfit 
the training corpus. However, this problem still exists
and requires further investigation.

\subsection{Exploration of Unlabeled Data}
\label{sec:unlabled_data}
To further explore the potential of leveraging self-supervised pre-training to solve STR
from a data perspective, we introduce a ViT-based model, namely MAERec.

\begin{figure}[t]
  \centering
  \includegraphics[scale=0.38]{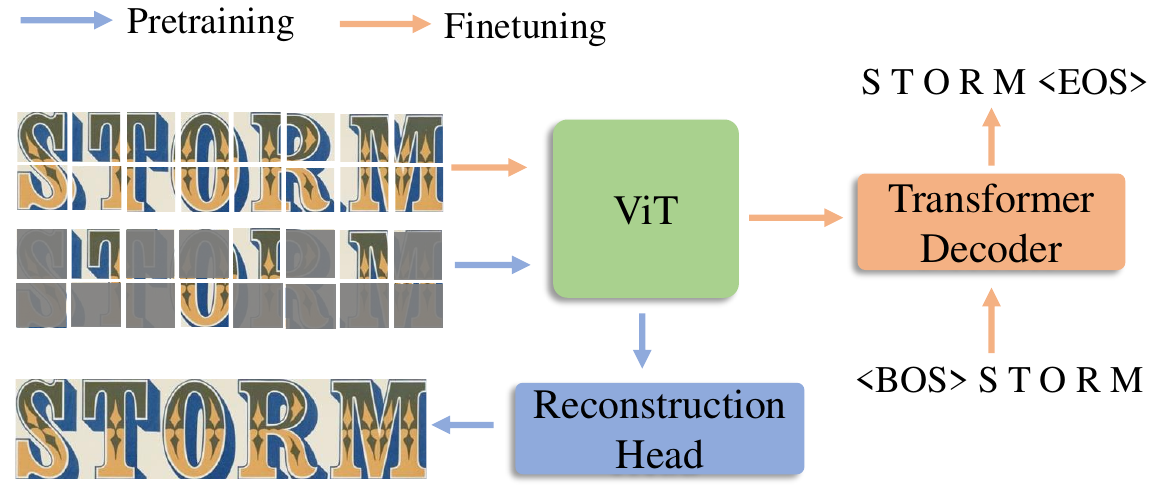}
  \caption{An overview of MAERec. It consists of a
    ViT \cite{dosovitskiy2020image} as the backbone and an auto-regressive Transformer decoder \cite{lee2020recognizing}.}
  \label{fig:5}
  \vspace{-0.5em}
\end{figure}

\textbf{Architecture of MAERec. }In Fig. \ref{fig:5}, we show the brief
architecture of MAERec. We choose Vision Transformer (ViT) \cite{dosovitskiy2020image} as the default backbone for its effortless applicability in masked image modeling \cite{he2022masked}. The input image is first fed into the ViT backbone
with a patch size of $4 \times 4$. The output sequence is  then passed to a Transformer decoder used in SATRN \cite{lee2020recognizing} for auto-regressive decoding to generate the predicted text. Details are in Appendix \ref{maerec}.

\begin{figure}[t]
  \centering
  \includegraphics[scale=0.38]{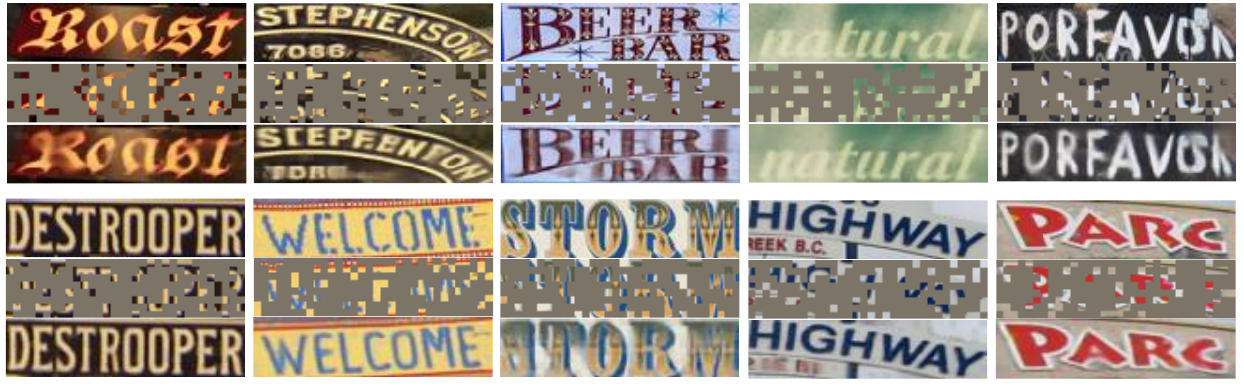}
  \caption{Reconstruction results on \textit{Union14M-L} images (images that are not used during
    pre-training). For each triplet, we show the
    ground truth (top), the masked image (middle), and the reconstructed image (bottom). The mask
    ratio is 75\%.}
  \label{fig:6}
  \vspace{-0.5em}
\end{figure}

\textbf{Pre-training. }To utilize the 10M unlabeled images in Union14M-U, we
pre-train the ViT backbone in MAERec through a masked image modeling task. We adopt the framework of MAE \cite{he2022masked} with minor modifications. The reconstruction
results are shown in Fig. \ref{fig:6}. The ViT backbone pre-trained on Union14-U can
yield convincing reconstructed text images, despite the mask ratio being high up to 75\%.
This indicates that the pre-trained ViT backbone can effectively capture the text structure in text
images and can learn useful textual representations.

\textbf{Fine-tuning. }After pre-training, we initialize MAERec with
the pre-trained ViT weight and fine-tune the whole model on Union14M-L. The results are shown in
Tab. \ref{tab:6} (last four rows). The performance of MAERec can be substantially improved
after pre-training, with an average accuracy gain
of 1.0\% on common benchmarks and 5.1\% on the Union14M-Benchmark, when using ViT-Small as the backbone.
Moreover, when scaling the backbone to ViT-Base, we can observe
significant performance improvements and MAERec can achieve an average accuracy
of 85.2\% on Union14M-Benchmark. This promising result demonstrates that utilizing massive unlabeled data can substantially improve the performance of STR models in real-world scenarios, and it is worth further exploration.

\textbf{Comparison with SOTA SSL methods. }We also compare our proposed MAERec with other self-supervised learning-based methods in STR, as shown in Table \ref{tab:7}. Despite the varying amounts of data used by different methods, MAERec outperforms its counterparts with a smaller data scale. It is noteworthy that MAERec utilizes a similar fine-tuning architecture with DiG \cite{yang2022reading} and a simpler pre-training framework, yet still achieves better results. This implies that the selection of data plays an even more critical role in self-supervised pre-training and fine-tuning.

\begin{table}[!t]
  \centering
  \caption{Comparison between MAERec and other self-supervised learning-based STR models with
    different pre-training and fine-tuning data. R stands for real data, and S stands
    for synthetic data. We report the average accuracy on six common benchmarks.}
  \resizebox{0.8\columnwidth}{!}{%
    \begin{tabular}{cccc}
    
      \hline
      Method                         & Pre-train       & Fine-tune & Avg.              \\ \hline
      PerSec \cite{liu2022perceiving} & 100M R         & 17M S    & 82.2             \\
      MaskOCR \cite{lyu2022maskocr}   & 4.2M R, 100M S & 17M S    & 92.6             \\
      DiG-S \cite{yang2022reading}    & 15.8M R, 17M S & 2.8M R   & 94.6             \\
      DiG-B \cite{yang2022reading}    & 15.8M R, 17M S & 2.8M R   & 95.0             \\ \hline
      MAERec-S (ours)                        & 10.6M R        & 3.2M R   & 95.1    \\
      MAERec-B (ours)                        & 10.6M R        & 3.2M R   & \textbf{96.2}    \\ \hline

    \end{tabular}}
  \label{tab:7}
  \vspace{-1.5em}
\end{table}

\section{Conclusion}
In this paper, we revisit scene text recognition from a data perspective. Despite the current benchmarks being close to saturation, we argue that the problem of STR remains unsolved, especially in real-world scenarios where current models struggle with numerous challenges. To explore the challenges that STR models still face, we consolidate a large-scale STR dataset for analysis and identified seven open challenges. Furthermore, we propose a challenge-driven benchmark to facilitate the future development of STR. Additionally, we reveal that the utilization of massive unlabeled data through self-supervised pre-training can remarkably enhance the performance of the STR model in real-world scenarios, suggesting a practical solution for STR from a data perspective. We hope this work can spark future research beyond the realm of existing data paradigms.

\section*{Acknowledgments}
This research is supported in part by NSFC (Grant No.: 61936003), National Key Research and Development Program of China  (2022YFC3301703), Zhuhai Industry Core and Key Technology Research Project (no. 2220004002350).

 {\small
    \bibliographystyle{ieee_fullname}
    \bibliography{paper}
  }
  
\newpage

\appendix
\section*{Appendix}
\section{Unrecognized Samples in Common Benchmarks}
\label{unrecog_sample}
\begin{figure*}[t]
  \centering
  \includegraphics[scale=0.43]{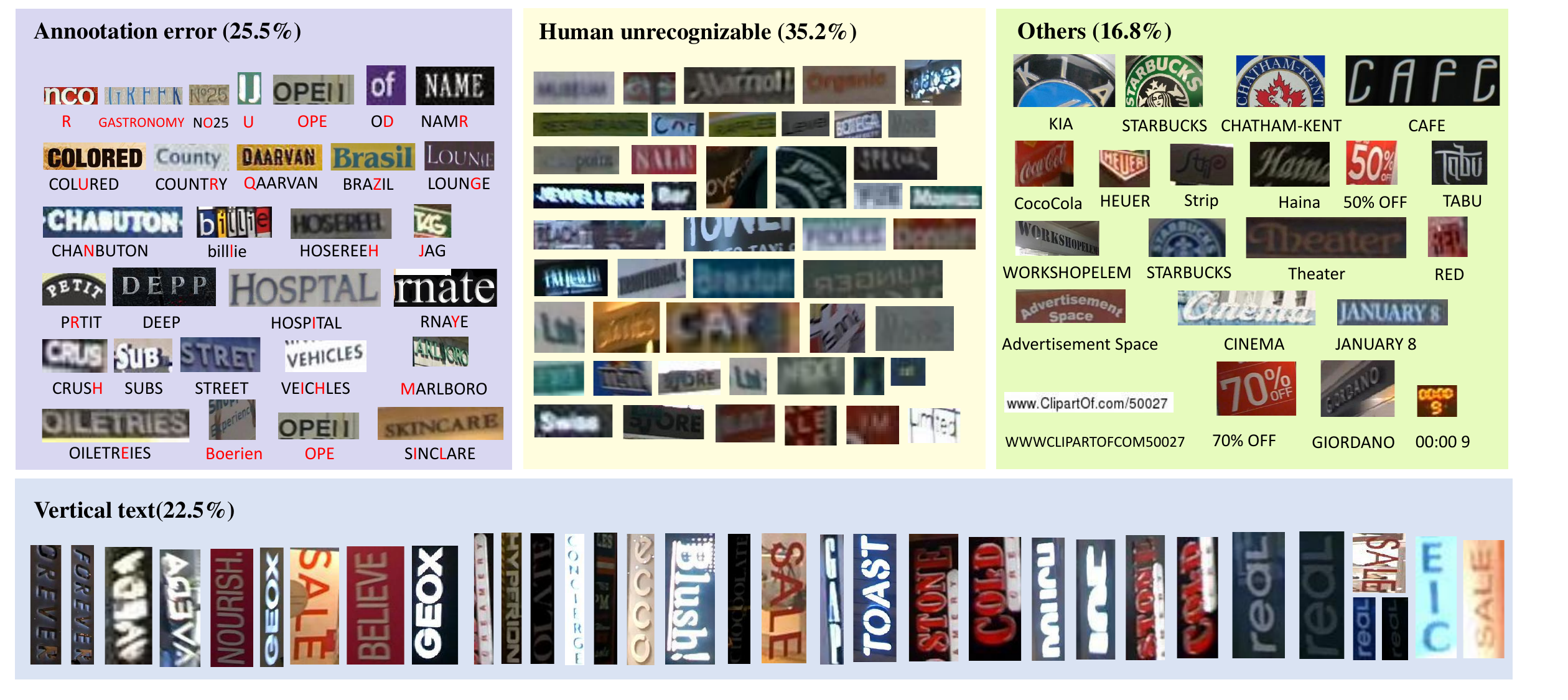}
  \caption{Examples of unrecognized samples in six common benchmarks.}
  \label{fig:8}
  \vspace{-0.5em}
\end{figure*}

In Fig. \ref{fig:8}, we show four types of images in the six common benchmarks that are
not correctly recognized by the ensemble of 13 STR models. Specifically, for human unrecognizable
images, we adopt the following criteria for adjudication: We recruit five human
experts, and each of them submits three possible predictions for each text image.
If all five experts failed to recognize a text image (i.e., 15 predictions in
total are incorrect), we regard it as a human unrecognizable sample. The majority
of these human unrecognizable samples exhibit high levels of blurriness and low
resolution. Furthermore, upon further examination of the 16.8\% of samples
that are classified as ``other", we can observe that many of them fall under the
categories of the seven challenges that we have discussed before, such as curve text,
multi-words text, and artistic text.

\section{More Details of Union14M}
\subsection{Construction of Union14M-U}
\label{construct_union14m}
\begin{figure}[t]
  \centering
  \includegraphics[scale=0.43]{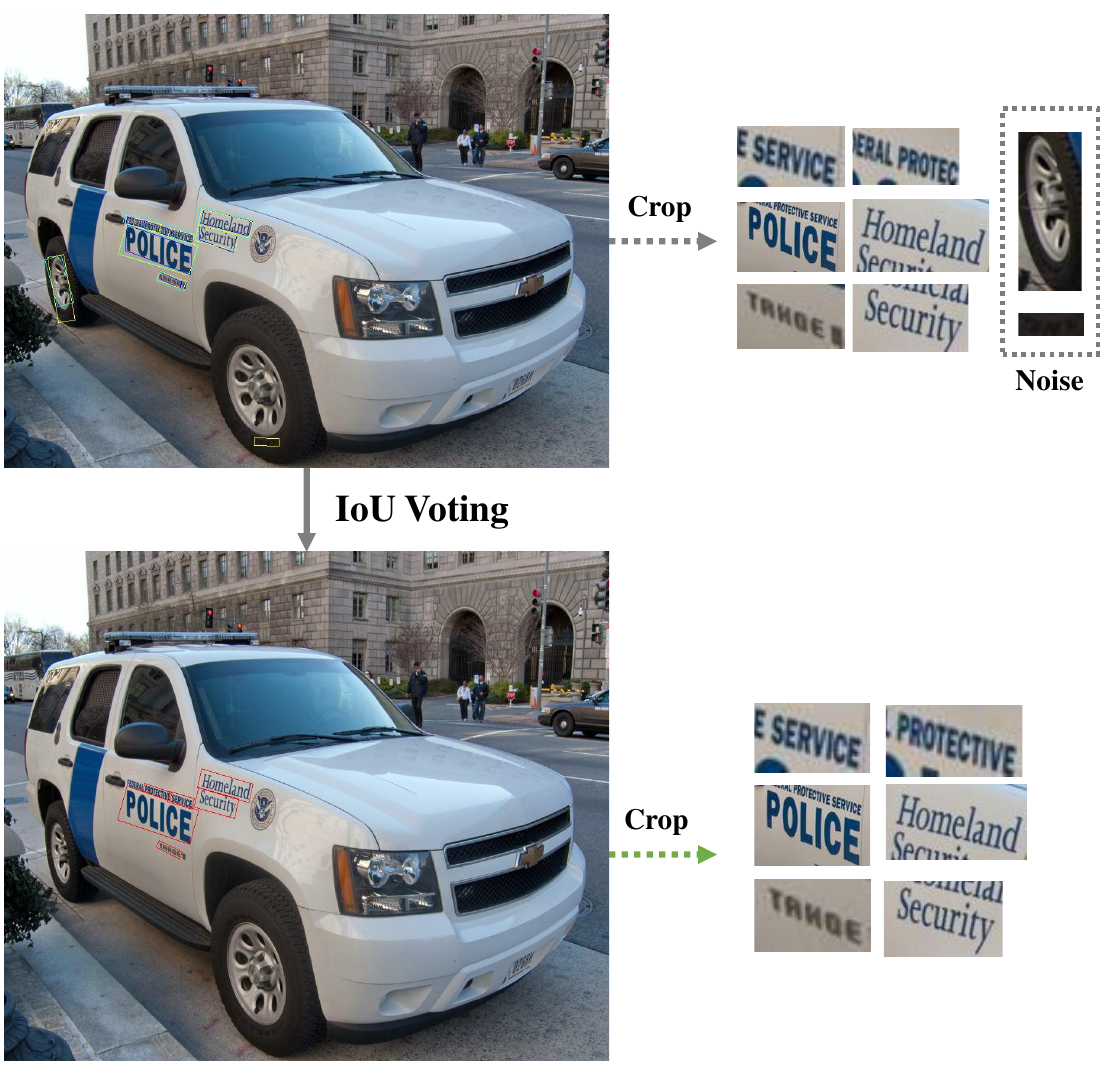}
  \caption{An illustration of our IoU voting strategy for collecting text instances.}
  \label{fig:7}
  \vspace{-0.5em}
\end{figure}

In order to gather a vast number of high-quality unlabeled text images, we utilize
three scene text detectors: DBNet++\footnote{\url{https://github.com/open-mmlab/mmocr/tree/main/configs/textdet/dbnetpp}} \cite{liao2022real},
BDN\footnote{\url{https://github.com/Yuliang-Liu/Box_Discretization_Network}} \cite{liu2019omnidirectional},
and EAST\footnote{\url{https://github.com/SakuraRiven/EAST}} \cite{zhou2017east}. We apply these
detectors to three large datasets: Book32\cite{iwana2016judging},
OpenImages\cite{kuznetsova2020open}, and Conceptual Captions (CC)\cite{sharma2018conceptual}.
However, directly using the results of these detectors is suboptimal due to the
presence of many false positive results produced by different detectors (e.g., in
Fig. \ref{fig:7}, the rear tire of the police car is detected as a text region
by two detectors). While missing detections can be tolerated given a large
amount of data, false detections are undesirable as they may introduce noise for
subsequent self-supervised learning. To address this issue, we adopt a simple  
Intersection over Union (IoU)
voting strategy to filter out false detections. Specifically, we identify regions
where the detected polygons of the three detectors have an IoU larger than 0.7
with respect to each other, and then we use the minimum axis-aligned rectangle of the
three detected polygons as the final prediction. Additionally, when selecting images from
OpenImages to construct Union14M-U, we exclude images with the same image ID in HierText \cite{long2022towards},
TextOCR \cite{singh2021textocr}, and InterOCR \cite{krylov2021open} since they
have already been used in Union14M-L. Using this
strategy, we obtain 10.6 million high-quality text instances in Union14M-U. 
It is noteworthy that all three detectors are trained on a singular dataset
(DBNet++ and EAST are trained on ICDAR2015 \cite{karatzas2015icdar}, BDN is
trained on MLT17 \cite{nayef2017icdar2017}),
which may contain inherent biases and lead to a lack of diversity in the
detected text instances. Therefore, investigating the usage of detectors
trained on larger datasets to obtain a larger number of text instances is a
potential direction for future research.

\begin{figure*}[h]
  \centering
  \includegraphics[scale=0.42]{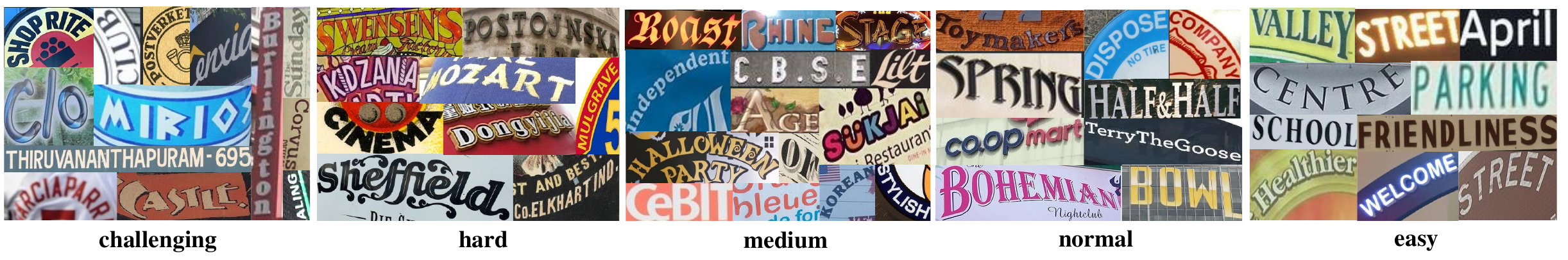}
  \caption{Examples of five difficulty levels in Union14M-L.}
  \label{fig:9}
  \vspace{-0.5em}
\end{figure*}

\begin{figure*}[t]
  \centering
  \includegraphics[scale=0.49]{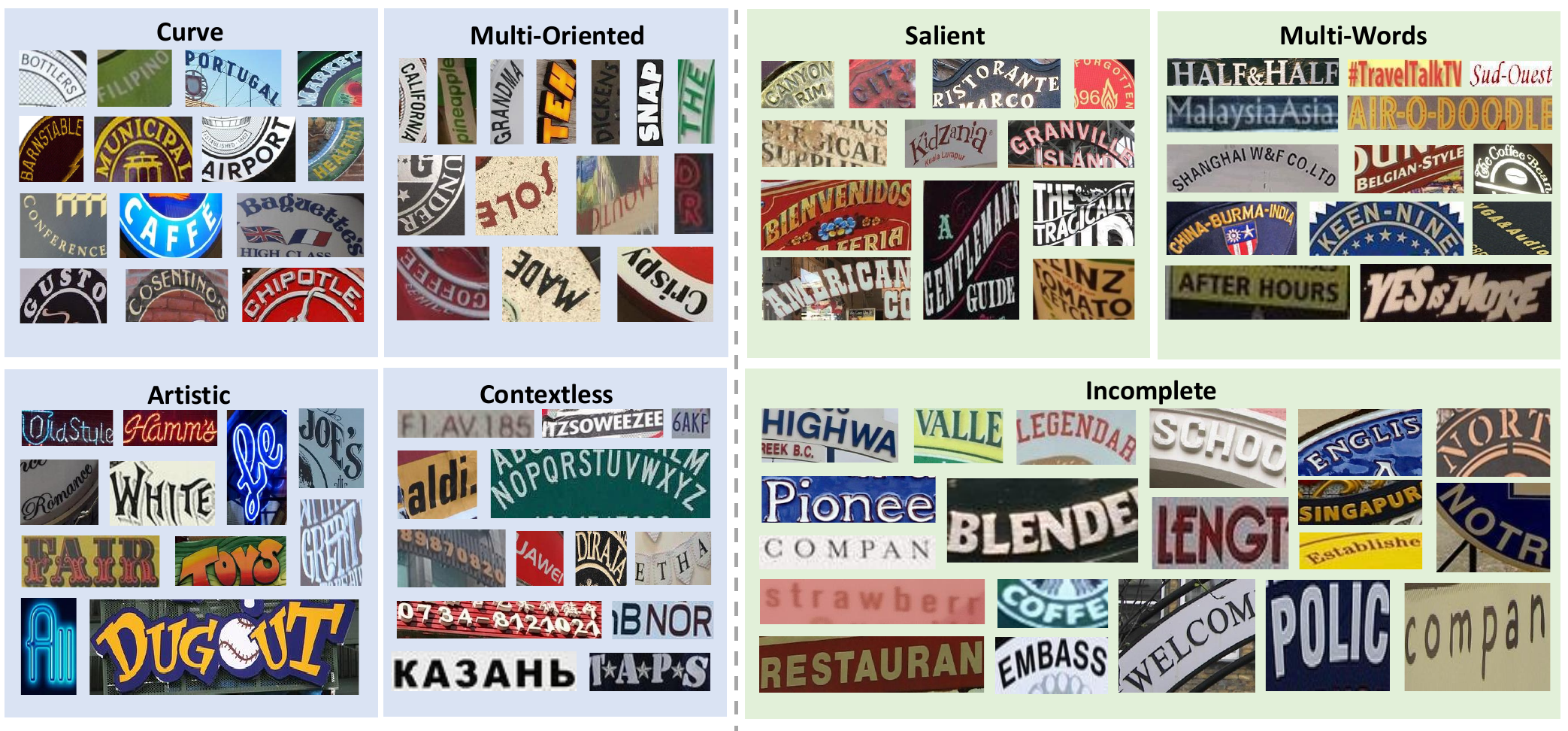}
  \caption{Examples of Union14M-Benchmark.}
  \label{fig:10}
  \vspace{-0.5em}
\end{figure*}

\begin{table}[h]
  \centering
  \caption{Comparison of different cropping ways. Settings remain the same as in Tab .\ref{tab:3}.}
  \resizebox{0.8\columnwidth}{!}{%
    \begin{tabular}{cccl}
      \hline
      Method                      & Training Data & Crop method       & Acc-UL                          \\ \hline
      SATRN \cite{lee2020recognizing}  & MJ, ST        & axis-aligned      & 72.09                           \\
      SATRN \cite{lee2020recognizing}                      & MJ, ST        & rotated           & 73.12  \\ \hline
      ABINet \cite{fang2021read}   & MJ, ST        & axis-aligned      & 70.73                           \\
      ABINet \cite{fang2021read}    & MJ, ST        & rotated           & 71.19  \\ \hline
    \end{tabular}}
  \label{tab:8}
\end{table}
\vspace{-0.5em}
\begin{table}[h]
  \centering
  \caption{Comparison of different cropping ways. Settings remain the same as in Tab .\ref{tab:6}.}
  \resizebox{0.8\columnwidth}{!}{%
    \begin{tabular}{cccl}
      \hline
      Method                      & Training Data     & Crop method       & Acc-CB      \\ \hline
      SATRN \cite{lee2020recognizing}   & Union14M-L        & axis-aligned      & 91.40       \\
      SATRN \cite{lee2020recognizing}      & Union14M-L        & rotated           & 89.03 (\textcolor{red}{-2.37})      \\ \hline
      ABINet \cite{fang2021read}     & Union14M-L        & axis-aligned      & 92.02       \\
      ABINet \cite{fang2021read}  & Union14M-L        & rotated           & 90.13 (\textcolor{red}{-1.89})      \\ \hline
    \end{tabular}}
  \label{tab:9}
\end{table}

\subsection{Comparison of Different Cropping Methods}
\label{crop}
We validate whether the large performance gap in Tab. \ref{tab:3} is caused by axis-aligned
crop. As shown in Tab .\ref{tab:8}, STR models still perform poorly when using rotated crop,
suggesting that the challenges inside Union14M-L are not caused by axis-aligned crops. Moreover, when training with rotated crop images, models exhibit inferior performance as shown in Tab .\ref{tab:9}, verifying our conjecture in that STR models will gain more robustness when training with a more noised text image. The inconsistency between STR and STD has been a less explored problem (E.g., The STR community used to focus on curve text recognition despite arbitrary shape text detectors being famous).

\subsection{Difficulty Assignment in Union14M-L}
\label{assignment}
\begin{figure}[t]
  \centering
  \includegraphics[scale=0.35]{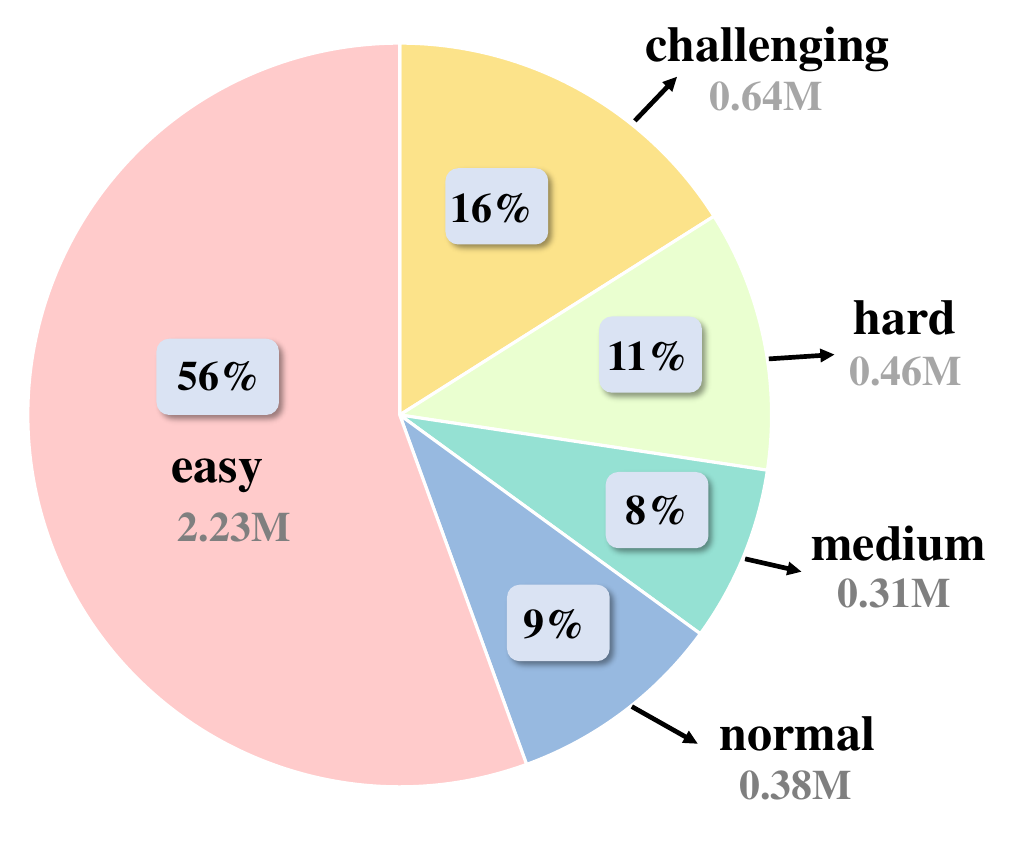}
  \caption{The proportion of samples with different difficulty levels in Union14M-L.}
  \label{fig:11}
  \vspace{-2.0em}
\end{figure}
Our focus is on analyzing the challenges that existing STR models encounter in
real-world scenarios. Therefore, we are interested in analyzing the samples
that present difficulties. As shown in Fig. \ref{fig:9}, we categorize the images
in Union14M-L into five difficulty levels using an error voting method.
Specifically, given an image $I$ and its corresponding ground
truth $Y$, we conduct forward inference on $I$ using the 13 STR models,
and the prediction results are denoted as $[Y_1, Y_2, \cdots, Y_{13}]$.
The voting list is defined as $V = [v_1, v_2, \cdots, v_{13}]$, where $v_i$ is defined
as:
\begin{equation}
  v_i = \begin{cases}
    1, & \text{if } Y_i = Y \\
    0, & \text{otherwise}
  \end{cases}
\end{equation}
Then each image is empirically assigned to a difficulty level according to the number of correct
predictions:
\begin{equation}
  \text{level} = \begin{cases}
    challenging,      & \text{if sum(V)} = 0          \\
    hard, & \text{if sum(V)} \in [1, 4]   \\
    medium,      & \text{if sum(V)} \in [5, 7]   \\
    normal,    & \text{if sum(V)} \in [8, 10]  \\
    easy,    & \text{if sum(V)} \in [11, 13] \\
  \end{cases}
\end{equation}
The subsets exhibit distinct characteristics based on their respective difficulty levels. For instance, the challenging set comprises a substantial number of images containing curve and vertical text, while the easy set primarily features clear samples and a clear background. The proportion of the images in each difficulty level is illustrated in Fig. \ref{fig:11}

\subsection{Consolidation of Union14M-Benchmark}

In this section, we provide more information on how we consolidate the
Union14M-Benchmark. For each of the seven challenges, excluding incomplete
text, we initially collect several reference images from Union14M-L that aligned with the definition of each of the seven challenges. We then recruit five human
experts to identify candidate images that shared similarities with the reference images. Subsequently, we manually examined each candidate image and eliminated images that did not meet the specified challenge criteria. Additionally, we also thoroughly recheck the annotations of all images, including digits, cases, and symbols to ensure the quality of the benchmark. For the incomplete text subset, all 1495 images are randomly sampled from the easy set of Union14M-L, and we cropped the first or last letter of each text image.

For the general subset, we sample 20\% of the images from each of the five difficulty
levels evenly to form the general subset with 400,000 images. With such uniform
sampling, the images in the general subset will be more uniformly distributed and
more representative. Since the sampling is random, the general subset
may have some annotation errors and human unrecognizable samples, as in the six common
benchmarks. However, due to a large amount of data, it will take much manual
effort to correct these errors, and we also hope that the academic community can work together to
correct the errors. In Fig. \ref{fig:10}, we show more samples of Union14M-Benchmark.

\begin{table}[t]
  \caption{Vision Transformer variants used in MAERec.}
  \vspace{+0.5em}
  \resizebox{\columnwidth}{!}{%
  \begin{tabular}{ccccc}
     \hline
     Model     & Layers & Hidden size & MLP size & Heads \\ \hline
     ViT-Small & 12     & 384         & 1536     & 6     \\
     ViT-Base  & 12     & 768         & 3072     & 12    \\ \hline
     \end{tabular}
  } 
  \label{tab:10}
  \vspace{-0.5em}
\end{table}

\section{Inplementation Details of MAERec}
\label{maerec}
\subsection{Vision Transformer}
\label{sec:3.1}
We use vallina Vision Transformer (ViT) \cite{dosovitskiy2020image}
as the backbone of MAERec, since it can be easily adapted to masked-image-modeling
pre-training. A ViT is composed of a patch embedding layer, position embedding,
and a sequence of Transformer blocks.

\textbf{Patch Embedding:} Since a ViT takes a sequence as input, the patch embedding
layer is used to convert the input image into a sequence of patches. Specifically,
given a text image of size $x \in \mathbb{R}^{H \times W \times C}$, we first
resize it to $x_{r} \in \mathbb{R}^{H_{r} \times W_{r} \times C}$,
where $H_{r}=32$ and $W_{r}=128$ following the common practice in
STR. We then use a patch embedding layer with a patch size of $4 \times 4$ to
split the image into non-overlapping patches, in this case, there are 256 patches
in total. Each patch is linearly projected to a $d$-dimensional vector, where $d$
is the embedding dimension of the patch embedding layer. 

\textbf{Position Embedding:} To retain positional information in the image, patch
embeddings are added with positional embeddings. Specifically, we use sinusoidal positional
embeddings in the original ViT \cite{dosovitskiy2020image} as follows:
\begin{equation}
  \begin{split}
  \text{PosEnc}(pos, 2i) &= \sin\left(\frac{pos}{10000^{2i/d}}\right) \\
  \text{PosEnc}(pos, 2i+1) &= \cos\left(\frac{pos}{10000^{2i/d}}\right)
  \end{split}
\end{equation}

where $\text{PosEnc}(pos, 2i)$ and $\text{PosEnc}(pos, 2i+1)$ represent the
$2i$-th and $(2i+1)$-th dimensions of the positional encoding for a given
position $pos$. $d$ represents the embedding dimension, and $i$ ranges from
0 to $\lfloor(d/2)\rfloor - 1$.

\textbf{Transformer blocks:} A Transformer block consists of alternating
layers of multi-head self-attention (MHSA) and MLP blocks. Given an input
sequence of embeddings $X \in \mathbb{R}^{L \times d}$, where $L$ is the sequence
length and $d$ is the embedding dimension, the transformer block can be
computed as follows:
\begin{equation}
\text{Block}(X) = \text{LN}(X + \text{LN}(\text{FFN}(\text{LN}(\text{MHSA}(X)))))
\end{equation}

where $\text{LN}$ is the layer normalization layer, $\text{FFN}$ is the
feed-forward network, and $\text{MHSA}$ is the multi-head self-attention layer.
We show the configuration of the ViT variants used in MAERec in Table \ref{tab:1}.

\subsection{Masked Image Modeling Pre-training}
We adopt MAE \cite{he2022masked} framework to pre-train the ViT backbone in MAERec.

\textbf{Encoder in MAE. } We use ViT described in Section \ref{sec:3.1} as the
encoder in MAE. Specifically, given patches $x \in \mathbb{R}^{N \times d}$,
where $N$ is the number of patches and $d$ is the embedding dimension of the
patch embedding layer, we randomly mask 75\% of the input patches and only send
the remaining 25\% visible patches to the ViT encoder. The mask size is set to
$4 \times 4$ to be consistent with the patch size.

\textbf{Decoder in MAE. }The decoder in MAE is input with the full set of tokens
including patch-wise representations from the ViT encoder and learnable mask
tokens put in the positions of masked patches. By adding positional embeddings
to all the input tokens, the decoder is able to reconstruct the original image
from the masked patches. Specifically, we adopt the original decoder used in MAE,
which is 8 layers of Transformer blocks and a linear layer to reconstruct the
text images from input tokens. The embedding dimension of Transformer blocks
is 512 and the number of heads is set to 16. The expanding factor of the MLP layer
is set to 4. 

\textbf{Reconstruct target. }The decoder in MAE is trained to reconstruct the
normalized pixel values of the original image, supervised by MSE loss.

\textbf{Optimization. }We adapt AdamW \cite{loshchilov2018decoupled} optimizer to 
pre-train the model on the 10.6M images of Union14M-U for 20 epochs with an initial
learning rate of 1.5e-4. The cosine learning rate scheduler is used with 2
epochs of linear warm-up. The pre-training image size is set to 32$\times$128,
and we use no data augmentation. The batch size is set to 256. Pre-training is
conducted with 4 NVIDIA A6000 (48GB RAM) GPUs.

\subsection{Fine-tuning for Scene Text Recognition}

\textbf{Auto-Regressive Transformer decoder. }We use the Transformer decoder
in \cite{na2022multi} for its superior performance in scene text recognition
. Specifically, we use six layers of Transformer
decoder to predict text sequence in an auto-regressive manner. The embedding
dimension of the Transformer decoder is set to 384 and 768 for the small and
base models respectively. The number of heads is set to 8.

\textbf{Optimization. }To be consistent with the pre-training
process, we still employ the AdamW optimizer with a weight decay of 0.01, and the cosine
learning rate scheduler without warm-up to train the model for 10 epochs. The batch
size is set to 64, and the initial learning rate is set to 1e-4. We also adopt the
same data augmentation strategy in \cite{fang2021read}. Fine-tuning is conducted
with 4 NVIDIA 2080Ti (11GB RAM) GPUs.

\section{More Experiment Analysis}

\subsection{Sources of the 13 STR Models}
In Tab. \ref{tab:11}, we list the sources of the 13 publicly available STR models.

\begin{table}[t]
  \caption{The sources of the 13 publicly available STR models.}
  \resizebox{\columnwidth}{!}{%
  \begin{tabular}{c|c|c}
    \hline
    Method        & Link                                                                      & Official ? \\ \hline
    CRNN          & \url{https://github.com/Mountchicken/Text-Recognition-on-Cross-Domain-Datasets} & No         \\
    SVTR          & \url{https://github.com/PaddlePaddle/PaddleOCR}                                & Yes        \\
    MORAN         & \url{https://github.com/Canjie-Luo/MORAN\_v2}                                   & Yes        \\
    ASTER         & \url{https://github.com/Mountchicken/Text-Recognition-on-Cross-Domain-Datasets} & No         \\
    NRTR          & \url{https://github.com/open-mmlab/mmocr/tree/main}                             & No         \\
    SAR           & \url{https://github.com/open-mmlab/mmocr/tree/main}                             & No         \\
    DAN           & \url{https://github.com/Wang-Tianwei/Decoupled-attention-network}               & Yes        \\
    SATRN         & \url{https://github.com/open-mmlab/mmocr/tree/main}                             & No         \\
    RobustScanner & \url{https://github.com/open-mmlab/mmocr/tree/main}                             & Yes        \\
    SRN           & \url{https://github.com/PaddlePaddle/PaddleOCR}                                 & No         \\
    ABINet        & \url{https://github.com/open-mmlab/mmocr/tree/main}                             & No         \\
    VisionLAN     & \url{https://github.com/wangyuxin87/VisionLAN}                                  & Yes        \\
    MATRN         & \url{https://github.com/byeonghu-na/MATRN}                                      & Yes        \\ \hline
    \end{tabular}
  } 
  \label{tab:11}
  \vspace{-1.0em}
\end{table}

\begin{table*}[t]
  \renewcommand{\arraystretch}{1.22}
  \caption{Performance (WA) of models trained on the training set of \textbf{Union14M}. In
  WA and WAIC metrics, it is impractical to measure the performance of the model on
  incomplete text set, because the performance is affected by whether the model 
  can correctly predict the case and symbols. For instance, if the model is
  wrong in case prediction, it will be considered as a false prediction in WA metric, and the
  error of incomplete text will be ignored. }
  \resizebox{\linewidth}{!}{
   \begin{tabular}{ccccccccc|cccccccc}
      \hline
      \multirow{2}{*}{Type}                          & \multirow{2}{*}{Method} & \multicolumn{7}{c|}{Common Benchmarks}                                                                                                                                                                                                                                                                                                                    & \multicolumn{8}{c}{Union4M Benchmarks}                                                                                                                                         \\ \cline{3-17} 
                                                     &                         & \begin{tabular}[c]{@{}c@{}}IIIT\\ 3000\end{tabular} & \begin{tabular}[c]{@{}c@{}}IC13\\ 1015\end{tabular} & \begin{tabular}[c]{@{}c@{}}SVT\\ 647\end{tabular} & \begin{tabular}[c]{@{}c@{}}IC15\\ 2077\end{tabular} & \begin{tabular}[c]{@{}c@{}}SVTP\\ 645\end{tabular} & \begin{tabular}[c]{@{}c@{}}CUTE\\ 288\end{tabular} & Avg                       & Curve & \begin{tabular}[c]{@{}c@{}}Multi-\\ Oriented\end{tabular} & Artistic & Contextless & Salient & \begin{tabular}[c]{@{}c@{}}Multi-\\ Words\end{tabular} & General & Avg  \\ \hline
      \multirow{2}{*}{CTC}                           & CRNN \cite{7801919}                    & 48.0                                                & 44.4                                                & 60.9                                              & 68.2                                                & 70.4                                               & 78.5                                               & 61.7                      & 18.8  & 4.2                                                       & 28.3     & 37.9        & 14.4    & 21.4                                                   & 56.7    & 26.0 \\
                                                     & SVTR \cite{du2022svtr}                  & 50.5                                                & 46.4                                                & 66.3                                              & 79.9                                                & 61.7                                               & 89.9                                               & 65.8                      & 69.9  & 66.2                                                      & 45.1     & 61.9        & 66.4    & 40.9                                                   & 73.1    & 60.5 \\ \hline
      \multicolumn{1}{l}{\multirow{7}{*}{Attention}} & MORAN \cite{luo2019moran}                  & 50.2                                                & 45.4                                                & 63.5                                              & 75.2                                                & 59.2                                               & 85.8                                               & 63.2                      & 41.9  & 12.0                                                      & 39.3     & 49.7        & 39.4    & 35.5                                                   & 41.4    & 37.0 \\
      \multicolumn{1}{l}{}                           & ASTER \cite{shi2018aster}                   & 49.1                                                & 45.0                                                & 64.8                                              & 73.8                                                & 58.0                                               & 83.7                                               & 62.4                      & 36.9  & 12.1                                                      & 35.6     & 46.9        & 29.0    & 33.4                                                   & 62.9    & 36.7 \\
      \multicolumn{1}{l}{}                           & NRTR \cite{sheng2019nrtr}                   & 50.5                                                & 47.1                                                & 67.7                                              & 77.1                                                & 60.3                                               & 90.3                                               & 65.5                      & 47.3  & 38.6                                                      & 47.8     & 64.3        & 38.7    & 49.5                                                   & 71.4    & 51.1 \\
      \multicolumn{1}{l}{}                           & SAR \cite{li2019show}                     & 50.5                                                & 46.7                                                & 67.1                                              & 83.5                                                & 62.6                                               & 90.6                                               & 66.8                      & 66.1  & 53.4                                                      & 53.3     & 66.6        & 55.4    & 49.8                                                   & 72.1    & 59.5 \\
      \multicolumn{1}{l}{}                           & DAN \cite{lee2020recognizing}                     & 49.6                                                & 46.3                                                & 64.8                                              & 74.4                                                & 57.7                                               & 84.7                                               & 62.9                      & 43.9  & 21.9                                                      & 43.7     & 55.1        & 39.8    & 38.4                                                   & 65.1    & 44.0 \\
      \multicolumn{1}{l}{}                           & SATRN \cite{wang2020decoupled}                  & 50.7                                                & 47.3                                                & 69.4                                              & 83.5                                                & 65.0                                               & 93.4                                               & 68.8                      & 72.0  & 63.8                                                      & 58.9     & 69.5        & 67.6    & 45.8                                                   & 77.2    & 65.2 \\
      \multicolumn{1}{l}{}                           & RobustScanner \cite{yue2020robustscanner}          & 50.2                                                & 46.4                                                & 67.4                                              & 79.0                                                & 61.6                                               & 91.0                                               & 65.9                      & 63.3  & 51.0                                                      & 54.0     & 72.7        & 54.7    & 46.7                                                   & 71.9    & 59.2 \\ \hline
      \multirow{4}{*}{LM}                            & SRN \cite{yu2020towards}                    & 50.1                                                & 45.5                                                & 64.3                                              & 74.3                                                & 58.8                                               & 87.8                                               & 63.5                      & 48.0  & 19.3                                                      & 43.2     & 54.9        & 39.9    & 27.7                                                   & 42.9    & 39.4 \\
                                                     & ABINet \cite{fang2021read}                 & 50.5                                                & 47.0                                                & 69.2                                              & 83.5                                                & 65.6                                               & 90.6                                               & 67.7                      & 72.2  & 58.7                                                      & 57.4     & 66.0        & 67.6    & 41.5                                                   & 75.6    & 62.7 \\
                                                     & VisionLAN \cite{wang2021two}              & 50.4                                                & 45.8                                                & 66.0                                              & 75.6                                                & 60.3                                               & 90.6                                               & 64.8                      & 68.0  & 54.7                                                      & 50.1     & 58.8        & 62.5    & 36.9                                                   & 70.5    & 57.4 \\
                                                     & MATRN \cite{na2022multi}                  & 50.9                                                & 47.2                                                & 69.6                                              & 84.0                                               & 65.9                                               & 94.1                                               & 68.6                      & 78.4  & 65.0                                                      & 61.7     & 69.7        & 73.0    & 52.6                                                   & 76.6    & 68.1 \\ \hline
      \multirow{4}{*}{Ours}                          & MAERec-S w/o PT         & 51.0                                                & 47.7                                                & 68.6                                              & 82.6                                                & 64.7                                               & 93.4                                               & 68.0                      & 72.7  & 63.7                                                      & 57.7     & 70.4        & 67.9    & 48.6                                                   & 77.1    & 65.4 \\
                                                     & MAERec-S                & 51.0                                                & 47.7                                                & 69.4                                              & 82.9                                                & 66.8                                               & 94.1                                               & 68.7                      & 78.2  & 68.8                                                      & 63.7     & 76.5        & 73.2    & 50.1                                                   & 78.7    & 69.9 \\
                                                     & MAERec-B w/o PT         & \multicolumn{1}{l}{50.9}                            & \multicolumn{1}{l}{47.6}                            & \multicolumn{1}{l}{69.7}                          & \multicolumn{1}{l}{83.0}                            & \multicolumn{1}{l}{66.1}                           & 93.1                                               & \multicolumn{1}{l|}{68.4} & 73.7  & 65.2                                                      & 57.6     & 69.7        & 69.7    & 48.1                                                   & 78.1    & 66.0 \\
                                                     & MARec-B                 & 51.3                                                & 48.0                                                & 70.9                                              & 85.2                                                & 67.1                                               & 95.1                                               & 69.6                      & 85.3  & 81.4                                                      & 70.9     & 79.2        & 80.1    & 54.6                                                   & 82.1    & 76.2 \\ \hline                                               
      \end{tabular}}
  \label{tab:12}
  \vspace{-0.5em}
 \end{table*}

 \begin{table*}[t]
  \renewcommand{\arraystretch}{1.22}
  \caption{Performance (WAIC) of models trained on the training set of \textbf{Union14M}.}
  \resizebox{\linewidth}{!}{
    \begin{tabular}{ccccccccc|cccccccc}
      \hline
      \multirow{2}{*}{Type}                          & \multirow{2}{*}{Method} & \multicolumn{7}{c|}{Common Benchmarks}                                                                                                                                                                                                                                                                                               & \multicolumn{8}{c}{Union4M Benchmarks}                                                                                                                                         \\ \cline{3-17} 
                                                     &                         & \begin{tabular}[c]{@{}c@{}}IIIT\\ 3000\end{tabular} & \begin{tabular}[c]{@{}c@{}}IC13\\ 1015\end{tabular} & \begin{tabular}[c]{@{}c@{}}SVT\\ 647\end{tabular} & \begin{tabular}[c]{@{}c@{}}IC15\\ 2077\end{tabular} & \begin{tabular}[c]{@{}c@{}}SVTP\\ 645\end{tabular} & \begin{tabular}[c]{@{}c@{}}CUTE\\ 288\end{tabular} & Avg  & Curve & \begin{tabular}[c]{@{}c@{}}Multi-\\ Oriented\end{tabular} & Artistic & Contextless & Salient & \begin{tabular}[c]{@{}c@{}}Multi-\\ Words\end{tabular} & General & Avg  \\ \hline
      \multirow{2}{*}{CTC}                           & CRNN \cite{7801919}                   & 81.5                                                & 91.3                                                & 82.4                                              & 69.9                                                & 69.8                                               & 79.2                                               & 79.0 & 18.9  & 4.3                                                       & 31.9     & 39.3        & 15.1    & 21.5                                                   & 58.1    & 27.0 \\
                                                     & SVTR \cite{du2022svtr}                   & 85.8                                                & 94.7                                                & 92.4                                              & 82.1                                                & 85.1                                               & 91.0                                               & 88.5 & 70.5  & 66.6                                                      & 50.2     & 63.0        & 71.4    & 42.6                                                   & 74.7    & 62.7 \\ \hline
      \multicolumn{1}{l}{\multirow{7}{*}{Attention}} & MORAN \cite{luo2019moran}                  & 85.6                                                & 93.6                                                & 87.3                                              & 77.1                                                & 82.6                                               & 86.1                                               & 85.4 & 42.4  & 12.4                                                      & 44.3     & 51.1        & 41.0    & 36.8                                                   & 42.9    & 38.7 \\
      \multicolumn{1}{l}{}                           & ASTER \cite{shi2018aster}                  & 84.1                                                & 92.0                                                & 87.6                                              & 75.5                                                & 79.5                                               & 84.0                                               & 83.8 & 37.4  & 12.5                                                      & 39.2     & 47.9        & 30.2    & 34.5                                                   & 64.4    & 38.0 \\
      \multicolumn{1}{l}{}                           & NRTR \cite{sheng2019nrtr}                   & 85.7                                                & 96.2                                                & 92.3                                              & 78.8                                                & 83.9                                               & 90.3                                               & 87.9 & 47.9  & 39.1                                                      & 51.8     & 65.1        & 40.1    & 51.4                                                   & 72.9    & 52.6 \\
      \multicolumn{1}{l}{}                           & SAR \cite{li2019show}                     & 86.5                                                & 95.3                                                & 90.7                                              & 81.6                                                & 86.1                                               & 91.0                                               & 88.5 & 66.9  & 54.7                                                      & 58.0     & 69.0        & 57.0    & 51.2                                                   & 73.7    & 61.5 \\
      \multicolumn{1}{l}{}                           & DAN \cite{lee2020recognizing}                    & 84.8                                                & 94.6                                                & 86.7                                              & 76.6                                                & 78.5                                               & 84.7                                               & 84.3 & 44.6  & 22.1                                                      & 47.0     & 56.6        & 41.5    & 39.8                                                   & 66.7    & 45.5 \\
      \multicolumn{1}{l}{}                           & SATRN \cite{wang2020decoupled}                  & 86.6                                                & 96.2                                                & 93.5                                              & 85.5                                                & 89.9                                               & 93.4                                               & 90.9 & 73.0  & 64.7                                                      & 64.3     & 71.1        & 69.2    & 47.4                                                   & 78.8    & 66.7 \\
      \multicolumn{1}{l}{}                           & RobustScanner \cite{yue2020robustscanner}          & 85.8                                                & 95.1                                                & 90.4                                              & 80.8                                                & 85.6                                               & 92.0                                               & 88.3 & 64.2  & 52.8                                                      & 58.7     & 72.7        & 56.9    & 47.8                                                   & 73.5    & 60.9 \\ \hline
      \multirow{4}{*}{LM}                            & SRN \cite{yu2020towards}                    & 85.6                                                & 94.2                                                & 88.6                                              & 76.8                                                & 82.9                                               & 88.5                                               & 86.1 & 48.7  & 20.0                                                      & 47.6     & 57.9        & 41.6    & 27.9                                                   & 60.7    & 42.5 \\
                                                     & ABINet \cite{fang2021read}                 & 86.5                                                & 96.8                                                & 94.1                                              & 85.8                                                & 90.9                                               & 91.7                                               & 91.0 & 73.0  & 59.6                                                      & 62.2     & 66.3        & 69.5    & 43.1                                                   & 75.6    & 65.5 \\
                                                     & VisionLAN \cite{wang2021two}              & 86.1                                                & 94.6                                                & 89.3                                              & 82.1                                                & 84.3                                               & 91.3                                               & 88.0 & 68.8  & 55.2                                                      & 54.4     & 60.1        & 64.7    & 37.9                                                   & 72.1    & 57.4 \\
                                                     & MATRN \cite{na2022multi}                  & 87.0                                                & 97.1                                                & 94.4                                              & 86.3                                                & 92.1                                               & 94.4                                               & 91.9 & 79.3  & 66.0                                                      & 67.3     & 71.0        & 74.9    & 53.8                                                   & 78.4    & 70.0 \\ \hline
      \multirow{4}{*}{Ours}                          & MAERec-S w/o PT         & 86.8                                                & 96.9                                                & 93.7                                              & 84.9                                                & 89.6                                               & 93.8                                               & 91.0 & 73.7  & 64.4                                                      & 62.1     & 71.5        & 69.5    & 49.3                                                   & 78.7    & 67.0 \\
                                                     & MAERec-S                & 87.3                                                & 97.0                                                & 95.1                                              & 85.3                                                & 92.1                                               & 95.1                                               & 92.0 & 79.3  & 69.5                                                      & 68.9     & 77.8        & 75.1    & 51.9                                                   & 80.4    & 71.8 \\
                                                     & MAERec-B w/o PT         & 86.8                                                & 97.2                                                & 85.5                                              & 95.4                                                & 91.6                                               & 94.1                                               & 91.8 & 74.8  & 65.7                                                      & 62.1     & 80.0        & 71.6    & 50.2                                                   & 79.7    & 69.2 \\
                                                     & MARec-B                 & 87.9                                                & 97.8                                                & 96.5                                              & 87.7                                                & 93.8                                               & 95.8                                               & 93.2 & 86.6  & 82.1                                                      & 75.9     & 80.7        & 82.2    & 56.2                                                   & 83.8    & 78.2 \\ \hline                                                    
      \end{tabular}}
  \label{tab:13}
  \vspace{-0.5em}
 \end{table*}

\begin{figure*}[t]
  \centering
  \includegraphics[scale=0.38]{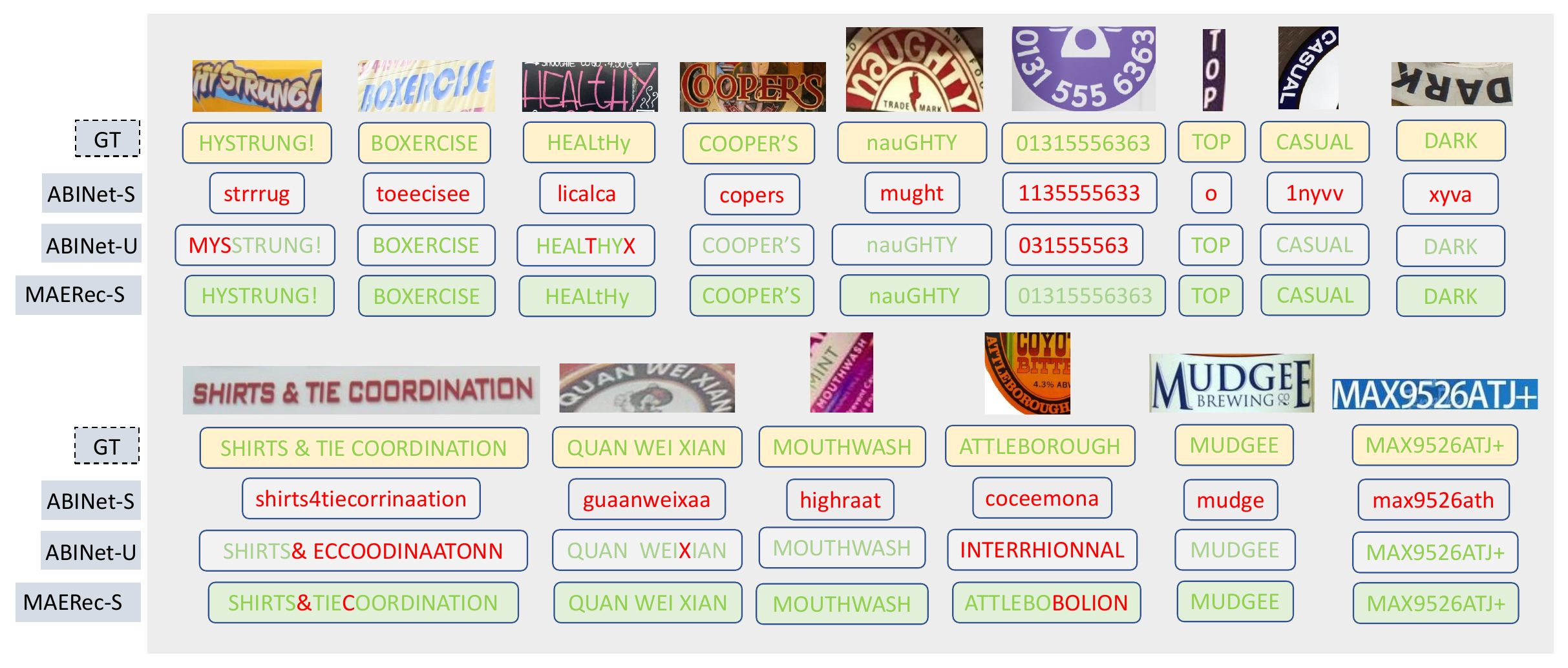}
  \caption{Recognition results on Union14M-Benchmark. GT stands for 
    ground truth. ABINet-S stands for ABINet\cite{fang2021read} trained on synthetic datasets (MJ and ST). 
    ABINet-U stands for ABINet trained on Union14M-L. The green text stands for correct recognition and the red text vice versa.}
  \label{fig:12}
\end{figure*}

\begin{figure}[t]
  \centering
  \includegraphics[scale=0.30]{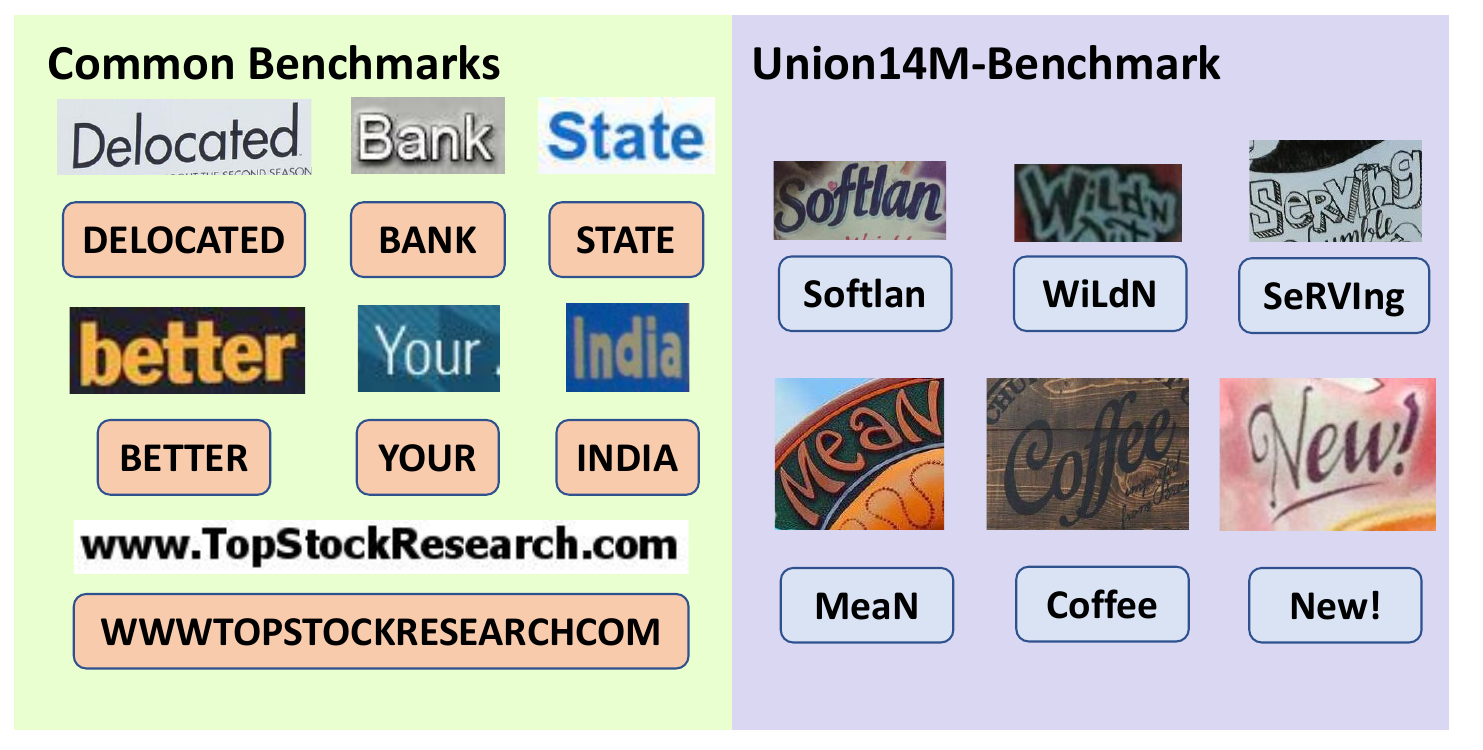}
  \caption{Compare the difference in the annotation of case between common benchmarks and Union14M-Benchmark.}
  \label{fig:13}
  \vspace{-1.5em}
\end{figure}

\begin{figure}[!h]
  \centering
  \includegraphics[scale=0.26]{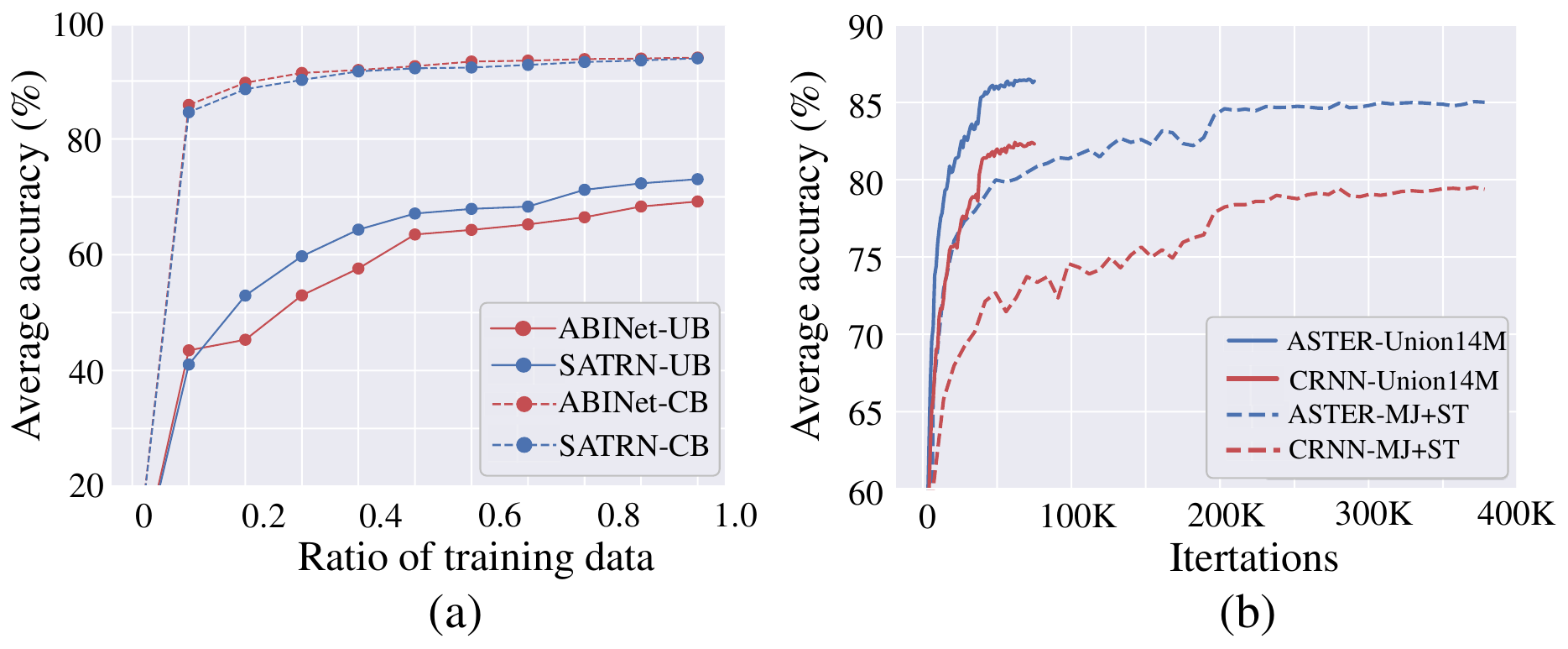}
  \caption{(a) Performance of models trained on increasing fractions of
    Union14M-L. CB denotes the six common benchmarks; UB denotes
    Union14M-Benchmark. (b) Performance evolution curves of
    models trained with Union14M-L or MJ \cite{jaderberg2016reading, jaderberg2014synthetic} and ST \cite{gupta2016synthetic} under the same configurations
    (number of epochs, optimizer, etc.), evaluated on the six common benchmarks.
  }
  \label{fig:14}
  \vspace{-1.0em}
\end{figure}

\subsection{WA and WAIC Metrics}
In Tab. \ref{tab:12} and Tab. \ref{tab:13}, we report the performance of models
trained with Union14M-L in terms of WA (word accuracy) and WAIC (word accuracy
ignore case) metrics, respectively. While most recent works evaluate
STR methods solely on the WAICS (word accuracy ignores case and symbols) metric, which ignores symbols and is
case-insensitive, some specific applications require the recognition of symbols
and cases, such as captcha recognition and license plate recognition. Compared to models
evaluated on the WAICS metric, we can observe a notable decrease in
performance when evaluated on both the WA and WAIC metrics. This phenomenon
can be attributed to the following reason:

\textbf{Incorrect case annotation. }The performance gap between WA and WAIC is
substantial in several common benchmarks, e.g., 50.3\% \textit{vs.} 85.89\% in
IIIT \cite{mishra2012scene} dataset (average accuracy of the 13 STR models). This is primarily
due to inconsistent case annotation. As shown in Fig. \ref{fig:13}, common benchmarks 
lack a unified annotation standard for the case. For example, in the IIIT dataset, the letters 
are all annotated in upper case, whereas in Union14M-Benchmark, we manually
check the case annotation of all the 9383 samples in challenge-specific subsets, and
correct any case errors. Therefore, the performance gap between WA
and WAIC metric in Union14M-Benchmark is much smaller (55.5\% \textit{vs.} 57.4\%).

\textbf{Lack of symbols. }Additionally, we note that there exists a performance gap between WAIC and WAICS
for STR models (88.3\% \textit{v.s} 91.2\% in common benchmarks; 57.4\% \textit{v.s}
62.7\% in Union14M-Benchmark). We suggest that this may be due to the
infrequent appearance of symbols in the training set in comparison to letters
and digits. This can be interpreted as a class imbalance issue, which requires
further investigation.

\subsection{Data Saturation}
We conducted a data ablation study to demonstrate the sufficiency of data in Union14M-L.
We select ABINet\cite{fang2021read} and
SATRN\cite{lee2020recognizing}, and train them on the
increasing fractions of the Union14M-L dataset. As depicted in Fig. \ref{fig:14}a, the accuracy
increases sharply in the beginning and eventually levels out. This indicates that
the real data in Union14M-L are sufficient, and adding more real
data may not lead to significant performance gain. Moreover, as shown in Fig. \ref{fig:14}b,
even though the data in Union14M-L are only 1/4 of the synthetic data, 
training on Union14M-L requires much fewer iterations (four times less)
to achieve higher accuracy, which aligns with the Green AI\cite{schwartz2020green} philosophy.

\subsection{Data Matters in Self-Supervised Pretraining }In Tab. \ref{tab:14}, we compare
different dataset combinations used in pre-training and fine-tuning. When pre-training
and fine-tuning are both performed on synthetic datasets, MAERec can barely
gain a performance boost (89.9 $\rightarrow$ 89.9 for CB, 46.0\% $\rightarrow$ 46.1\% for UB). However, when
fine-tuning is performed on Union14M-L, MAERec can exhibit a performance boost
when either pre-trained on synthetic datasets (73.5\% $\rightarrow$ 75.0\% for UB) or on
Union14M-U (73.5\% $\rightarrow$ 78.6\% for UB). This indicates that fine-tuning on 
real data is vital for self-supervised learning, and Union14M-U is
preferable to synthetic datasets for pre-training (78.6\% \textit{vs.} 75.0\%). 

\begin{table}[t]
  \centering
  \caption{Compare the performance of MAERec-S with different pre-training and
  fine-tuning datasets. Acc-CB denotes the average accuracy on six common benchmarks. Acc-UB
  denotes the average accuracy on Union14M-Benchmark (Exclude incomplete text subset).}
  \resizebox{0.74\columnwidth}{!}{%
  \begin{tabular}{ccccc}
    \hline
    No. & Pre-train  & Fine-tune & Acc-CB & Acc-UB \\ \hline
    1   & -          & MJ, ST     & 89.9  & 46.0  \\
    2   & -          & Union14M-L & 94.1  & 73.5  \\ \hline
    3   & MJ, ST     & MJ, ST     & 89.9  & 46.1  \\
    4   & MJ, ST     & Union14M-L & 94.0  & 75.0     \\
    5   & Union14M-U & Union14M-L & \textbf{95.1}  & \textbf{78.6}  \\ \hline
    \end{tabular}}
  \label{tab:14}
  \vspace{-1.0em}
\end{table}

\begin{figure*}[t]
  \centering
  \includegraphics[scale=0.36]{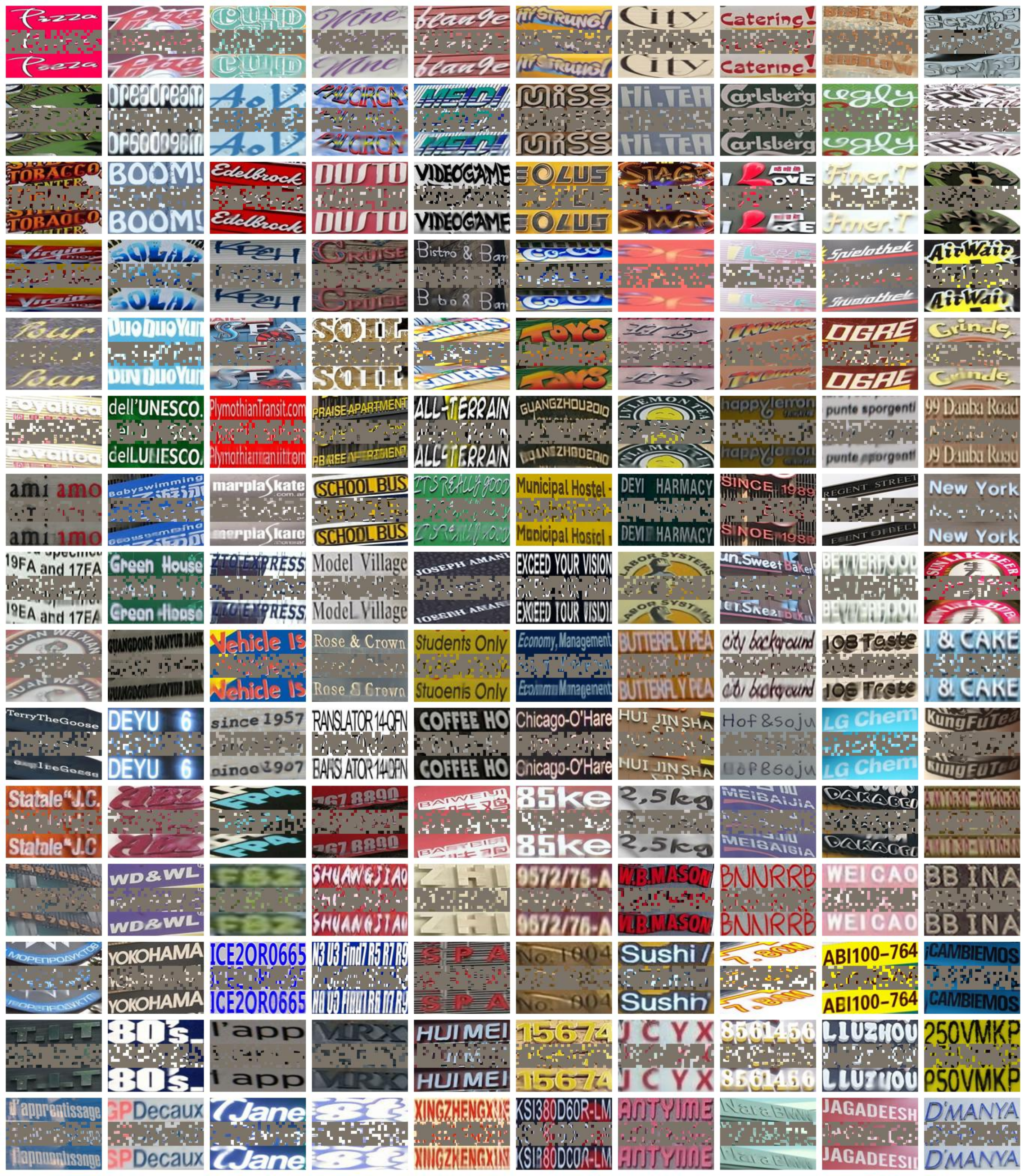}
  \caption{More reconstruction samples. For each triplet, we show the
  ground truth (top), the masked image (middle), and the reconstructed image (bottom).
  Images are from \textit{artistic text}, \textit{multi-words text}, and
  \textit{contextless text} in Union14M-Benchmark.
  }
  \label{fig:15}
  \vspace{+1.5em}
\end{figure*}

\subsection{Visualize Recognition Results}
We show some recognition results on Union14M-Benchmark in Fig. \ref{fig:12}.
Compared with models trained on synthetic data, training on Union14M can empower
STR models to cope with various complex real-world scenarios, thus significantly
improving their robustness.

\subsection{Why MIM Pre-training Works for STR}
When MAERec is pre-trained using MAE on Union14M-U, it shows significant improvement in the STR downstream task. The reason behind this improvement could be attributed to the pre-training process of MIM, where a large portion of the text image (75\%) is covered, resulting in only a few patches of each character being visible to the ViT backbone. As a result, if the decoder needs to reconstruct the original image, the ViT backbone must learn to recognize the smallest part of a character to infer the whole character, as shown in Fig. \ref{fig:15}. After pre-training, the ViT backbone has learned to differentiate between different characters during pre-training, and the downstream recognition task is essentially a classification task. Hence, the model's performance is significantly enhanced.

\end{document}